\documentclass[preprint,a4]{elsarticle}

\usepackage{graphicx}%
\usepackage{multirow}%
\usepackage{multicol}%
\usepackage{amsmath,amssymb,amsfonts}%
\usepackage{amsthm}%
\usepackage{mathrsfs}%
\usepackage[title]{appendix}%
\usepackage{xcolor}%
\usepackage{textcomp}%
\usepackage{manyfoot}%
\usepackage{booktabs}%
\usepackage{algorithm}%
\usepackage{algorithmicx}%
\usepackage{algpseudocode}%
\usepackage{listings}%
\usepackage{siunitx}
\usepackage{subcaption}
\usepackage{float}
\usepackage{geometry}%
\usepackage{hyperref}
\usepackage{cleveref}
\newgeometry{margin=1in}
\crefname{algocf}{alg.}{algs.}
\Crefname{algocf}{Algorithm}{Algorithms}

\usepackage{acro}
\DeclareAcronym{gdl}{
  short = GDL,
  long  = Graph Deep Learning,
  long-format = \emph ,
}
\DeclareAcronym{gnode}{
  short = GNODE,
  long  = Graph Neural ODE,
  long-format = \emph ,
}
\DeclareAcronym{gde}{
  short = GDE,
  long  = Graph Differential Equation,
  long-format = \emph ,
}
\DeclareAcronym{peml}{
  short = PEML,
  long = Physics-Enhanced Machine Learning,
  long-format = \emph ,
}
\DeclareAcronym{pgml}{
  short = PGML,
  long = Physics-Guided Machine Learning,
  long-format = \emph ,
}
\DeclareAcronym{piml}{
  short = PIML,
  long = Physics-Informed Machine Learning,
  long-format = \emph ,
}
\DeclareAcronym{kf}{
  short = KF,
  long = Kalman Filter,
  long-format = \emph ,
}
\DeclareAcronym{ekf}{
  short = EKF,
  long = Extended Kalman Filter,
  long-format = \emph ,
}
\DeclareAcronym{gnn}{
  short = GNN,
  long = Graph Neural Network,
  long-format = \emph ,
}
\DeclareAcronym{mpnn}{
  short = MPNN,
  long = Message-Passing Neural Network,
  long-format = \emph ,
}
\DeclareAcronym{gcn}{
  short = GCN,
  long = Graph-Convolutional Network,
  long-format = \emph ,
}
\DeclareAcronym{gkf}{
  short = GKF,
  long = Graph Kalman Filter,
  long-format = \emph ,
}
\DeclareAcronym{pggnode}{
  short = PiGGO,
  long = \textbf{P}hys\textbf{i}cs-\textbf{G}uided \textbf{G}raph Neural \textbf{O}DE,
  long-format = \emph ,
}
\DeclareAcronym{nmse}{
  short = NMSE,
  long = Normalised Mean Squared Error,
  long-format = \emph ,
}
\DeclareAcronym{shm}{
  short = SHM,
  long = Structural Health Monitoring,
  long-format = \emph ,
}
\DeclareAcronym{rnn}{
  short = RNN,
  long = Recurrent Neural Network,
  long-format = \emph ,
}

\makeatletter
\newcounter{phase}[algorithm]
\newlength{\phaserulewidth}

\makeatother

\makeatletter
\def\algbackskip{\hskip-\ALG@thistlm}
\makeatother

\begin{document}

\title{PiGGO: Physics-Guided Learnable Graph Kalman Filters for Virtual Sensing of Nonlinear Dynamic Structures under Uncertainty}


\author[1]{Marcus Haywood-Alexander\corref{cor1}}\ead{mhaywood@ethz.ch}

\author[1]{Gregory Duthé}\ead{dutheg@ethz.ch}

\author[1,2]{Eleni Chatzi}\ead{chatzie@ethz.ch}

\cortext[cor1]{Corresponding author}



\affiliation[1]{organization={Department of Civil, Environmental and Geomatic Engineering, ETH Zürich},
addressline={Wolfgang-Pauli Strasse 5},
postcode={8049},
city={Zürich},
country={Switzerland}}

\affiliation[2]{organization={Future Resilient Systems},
addressline={Singapore-ETH Centre},
city={Singapore},
country={Singapore}}


\begin{abstract}
Digital twins provide a powerful paradigm for diagnostic and prognostic tasks in the monitoring and control of engineered systems; however, their deployment for complex structures remains challenged by model-form uncertainty, arising from unknown nonlinear dynamics, and by sparse sensing. These limitations hinder reliable online state estimation using either purely physics-based or purely data-driven approaches. 

This work introduces the \emph{Physics-Guided Graph Neural ODE (PiGGO)} framework, a physics-informed, graph-based Bayesian state estimation approach in which a learned \emph{graph neural ordinary differential equation (GNODE)} serves as the continuous-time state-transition model within an \emph{extended Kalman filter}. The graph representation explicitly defines the system state-space, while physics-guided inductive biases encode known structural relationships and constrain the learning of nonlinear dynamics. 

By integrating graph-native learned dynamics with recursive Bayesian filtering, the proposed PiGGO framework enables online virtual sensing and uncertainty-aware state estimation for nonlinear systems with unknown model form, while maintaining generalisation across topologically similar structures. Numerical case studies demonstrate improved robustness to model uncertainty and measurement noise, outperforming both open-loop graph neural models and conventional filtering approaches in online prediction tasks.
\end{abstract}




\maketitle

\section{Introduction}\label{sec:intro}
The development of smart structures relies on digital twins that faithfully represent the dynamics of monitored structures \cite{michael2024integrating, torzoni2024digital}, enabling diagnostic and prognostic tasks within the context of \ac{shm} \cite{ye2019digital, lai2023digital, hielscher2023neural}. %
A primary function of digital twins is \emph{virtual sensing}, defined as the estimation of structural response, commonly represented by state variables $\mathbf{z}$, at unmeasured locations based on a sparse set of available observations $\mathbf{y}^*$ \cite{vettori2023adaptive}.
In many \emph{in-situ} monitoring scenarios, the accessible measurements are limited to accelerations, strains and external forces, while further quantities of primary interest — such as displacement and velocity — must be inferred indirectly \cite{vettori2020virtual}.  %
However, deploying such digital twins faces the challenge of model error -- both \emph{parameter} and \emph{model form} uncertainties \cite{kamariotis2024bayesian,kapteyn2021probabilistic, worden2020digital} -- and inherent operational variability that numerical models fail to capture \cite{simoen2015dealing,rios2020uncertainty, shi2024construction}. %
To maintain the fidelity required for reliable SHM, it is essential to ensure a digital twin architecture that is able to identify and compensate for both epistemic and aleatoric uncertainties \cite{worden2007fundamental}. %
In recent decades, this is often captured through the use of powerful data-driven techniques available from machine learning \cite{farrar2012structural}. %

Although linear systems are relatively simple to capture and model, real-world structures often exhibit complex nonlinearities as a result of material effects, discrete feedback forces, or joint mechanisms, which purely physics-based models struggle to accurately represent \cite{legault2012physical}. %
Conversely, purely data-driven models can bypass explicit modelling, but suffer under sparse measurements \cite{barthorpe2010model, ghadami2022data}. %
This motivates \ac{peml} approaches that combine physical structure with learning-based components \cite{cicirello2024physics,cross2021physics,haywood2024discussing}. %
A prominent class of PEML methods are \ac{piml} approaches, which leverage physics-based inductive biases to compensate for sensing limitations \cite{sun2024hybrid, zargar2024physics, zhang2024physics,haywood2025response,zhang2024dual}. %
However, these methods are restricted to \emph{offline} settings and rely on the assumption that the prescribed physics is sufficient \cite{gupta2025physics}. 
An alternative is learning a generative dynamics models, such as physics-guided methods, \cite{liu2022physics}, to capture system evolution directly. %

\begin{figure}[h!]
    \centering
    \includegraphics[width=0.9\textwidth]{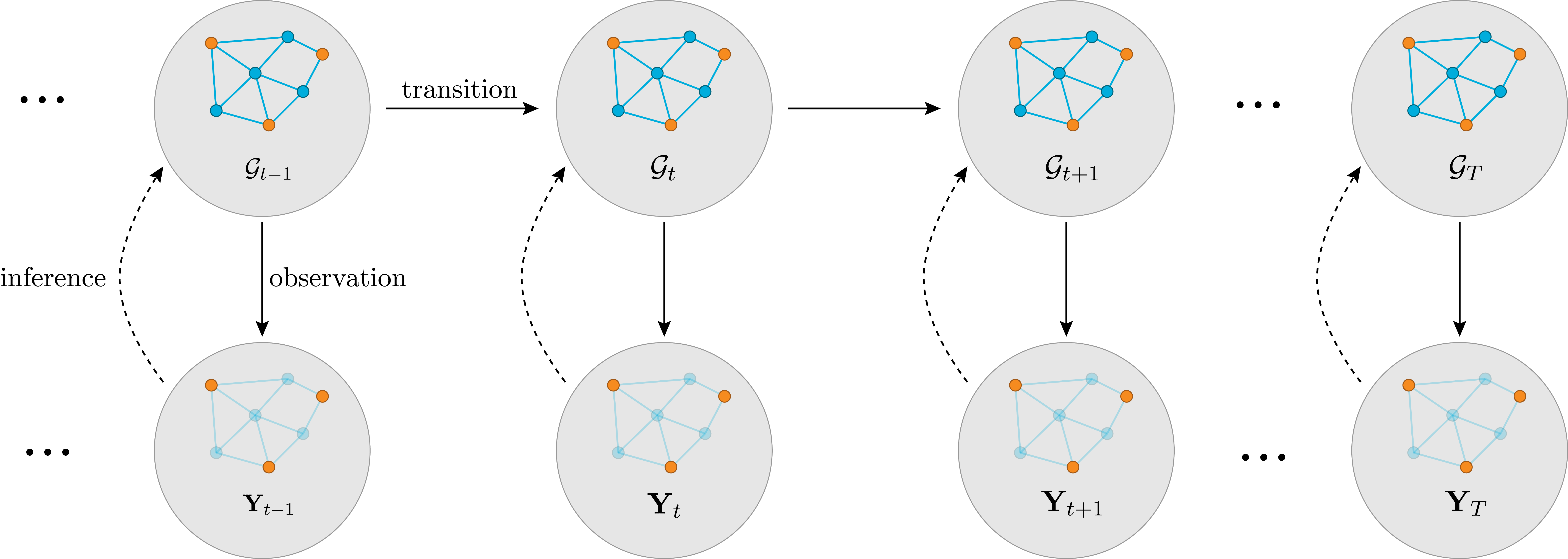}
    \caption{State-space transition model for virtual sensing on graphs.}
    \label{fig:ss-trans-model}
\end{figure}

For ML tasks where the underlying function to be learned is driven by a prescribable topology, \acp{gnn} offer a flexible, geometry-aware universal function approximator. %
The response of dynamic structures is driven by local interactive forces between components in the structure, therefore, \acp{gnn} provide a powerful framework for dynamical systems, using spatial inductive biases to improve observability, particularly across populations \cite{tsialiamanis2021foundations,jian2025using}. %
A subclass of \acp{gnn} is the \ac{gnode} \cite{poli2019graph}, which extends the \emph{Neural-ODE} \cite{chen2018neural} to graph-structured dynamical systems. %
The state transition is defined by a learnable graph-parameterised differential equation,
$\dot{\mathbf{H}} = \mathbf{F}(\mathcal{G})$ (cf. \Cref{fig:ss-trans-model}). %
State predictions are obtained by numerically integrating this vector field, while an observation operator maps the resulting continuous-time states to measurable outputs $\mathbf{Y}$. %
While \acp{gnode} function well for offline virtual sensing, in purely generative applications in an online setting, natural structural variability can lead to error accumulation and degraded performance when generalising across populations of structures \cite{tsialiamanis2023towards}. %

To mitigate this error accumulation, Bayesian filtering schemes are commonly employed \cite{tatsis2022sequential, azam2015dual}. %
While a powerful stochastic framework for virtual sensing of structural responses \cite{chui2009kalman, simon2006optimal}, the standard \ac{ekf} requires an accurate and differentiable state-transition model, often unavailable for complex nonlinear systems \cite{pyrhonen2023linearization, li2021parameter}. %
Recent work therefore integrates learnable models within the filtering loop to alleviate model bias and improve robustness \cite{liu2024neural, revach2022kalmannet, xia2024neural}. %

Kalman filters have recently been extended to graph-structured data \cite{li2023unscented}, however, as \acp{gkf} contain additional computational complexity, their methods of implementation vary depending on the problem context. %
In \cite{sagi2023extended}, the Kalman gain is calculated using the graph topology, thus simplifying the model formulation. %
Where adequate model form is available, but identification is difficult, \cite{sabbaqi2025gknetgraphkalmanfiltering}, uses \acp{gnn} to estimate the parameters of the prescribed underlying model. %
Alippi and Zambon (2023) exploited the recursive updating capabilities of Bayesian filtering to adaptively infer and update the graph topology, treating the adjacency matrix itself as an evolving latent variable \cite{alippi2023graph}. %
A common use of \acp{gkf} is in traffic prediction, where the capabilities of learned-model-based \ac{gkf} schemes for novelty detection are highlighted \cite{sun2022detecting}. %
The computational complexity of \acp{gnn} has limited the use of \acp{gkf} for high-dimensional data, and such filtering schemes are often applied to low-dimensional latent variables \cite{buchnik2024gsp}. %

In this work, we introduce the \emph{\ac{pggnode}} framework, a unified approach for learning and filtering nonlinear structural dynamics under uncertainty. %
The proposed methodology integrates a learned \ac{gnode} as a continuous-time, graph-structured state-transition model within an \ac{ekf}, thereby coupling physics-guided representation learning with recursive Bayesian state estimation. %
The \ac{pggnode} framework comprises three key components: (i) a graph-based representation that explicitly defines the system state-space, (ii) a physics-guided GNODE that captures the underlying nonlinear dynamics through inductive biases informed by structural mechanics, and (iii) an EKF that performs online state estimation and uncertainty quantification. %
By embedding structural relationships directly within the graph and constraining the learning process through physics-guided architectural design, the framework enables the identification of nonlinear dynamics beyond the reach of purely data-driven approaches. %

A key contribution of this work lies in the decoupled yet synergistic training and inference strategy: the \ac{pggnode} is first trained offline to learn a topology-aware, generalisable representation of the system dynamics, and is subsequently embedded within an EKF to enable online correction of model-form errors and measurement noise. %
This results in a scalable and robust virtual sensing framework capable of uncertainty-aware state estimation in nonlinear systems with unknown or partially known governing equations, while maintaining generalisation across topologically similar structures. %

The paper is organised as follows: Section~2 introduces the \ac{pggnode} framework for offline virtual sensing. Section~3 presents its integration within a graph Kalman filtering scheme for online inference. Section~4 evaluates the approach on nonlinear structural dynamics case studies, followed by conclusions and future research directions in Section~5.

\section{Proposed Method for Offline Virtual Sensing}\label{sec:off-virt-sens}

\subsection{Graph Neural Network}
We start by introducing the Graph Neural Network, as this acts as the basis for this approach, which relies on processing graph-structured data. %
At the core of the \ac{gnn} lies the \emph{message-passing} paradigm \cite{gilmer2017neural}, which provides a unified formulation for information exchange between nodes by aggregating features from their neighbourhoods. 
Architectures based on this principle are commonly referred to as \emph{message-passing neural networks} (MPNNs) and underpin most widely used GNN variants implemented in libraries such as \texttt{PyTorch-Geometric} \cite{fey2019fast}, \texttt{Deep Graph Library (DGL)} \cite{wang2019deep}, and \texttt{Spektral} \cite{grattarola2021graph}.

\begin{figure}[h]
	\centering
	  \includegraphics[width=0.75\textwidth]{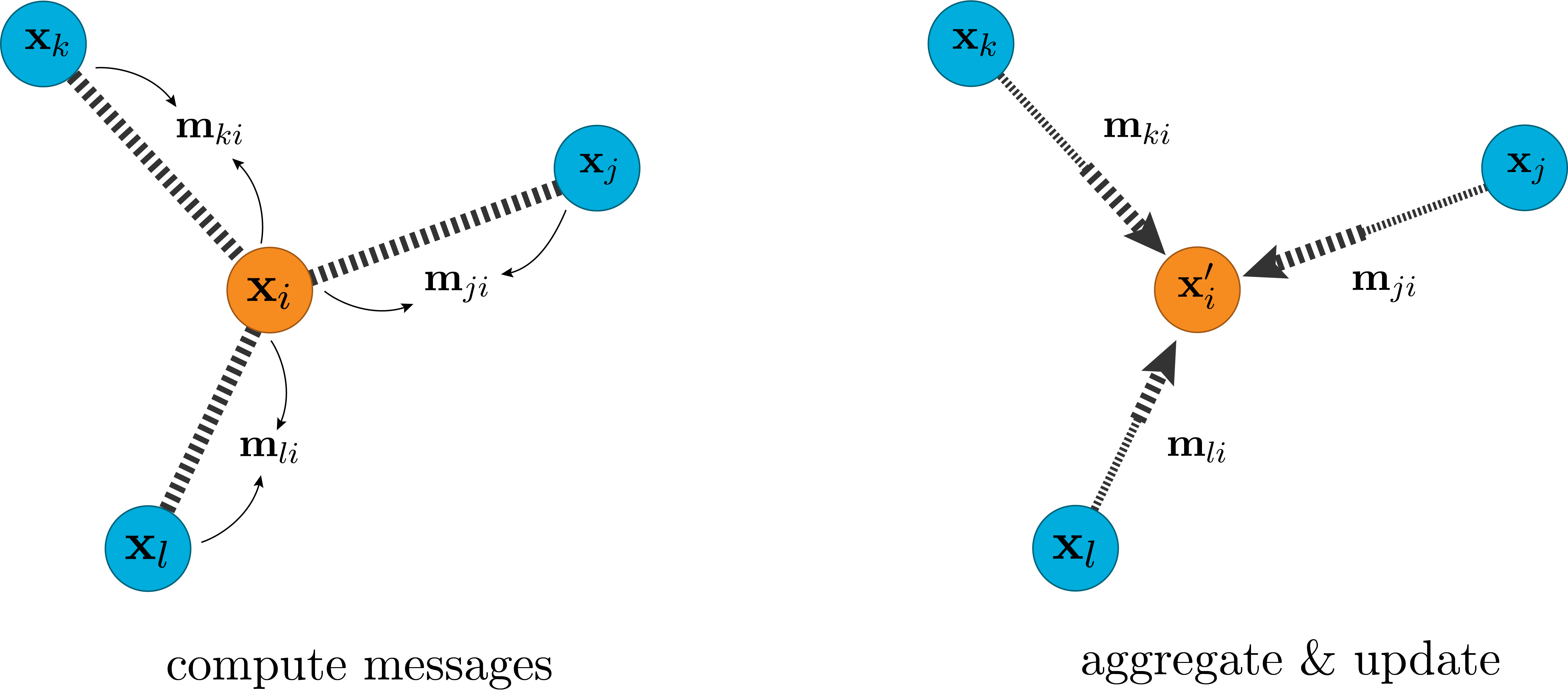}
        \caption{The two phases of message-passing in GNNs. Left: Initial message computation phase, where messages are computed on each edge. Right: Aggregation and update phase, where messages from neighbours are combined and the node's features are updated. The design choices for these two operations define different GNN variants.}
    \label{fig:message_passing}
\end{figure}

In the message-passing framework, information is propagated between connected nodes as follows.
For an edge from node $j$ to node $i$, a message $\mathbf{m}_{ji}$ is computed from the associated node and edge features,
\begin{equation}\label{eq:message}
\mathbf{m}_{ji} = \phi(\mathbf{x}_j, \mathbf{x}_i, \mathbf{e}_{ji}; \boldsymbol{\theta}_{\phi}),
\end{equation}
where $\mathbf{x}_i$ and $\mathbf{x}_j$ denote node features, $\mathbf{e}_{ji}$ edge features, and $\phi(\cdot)$ a learnable message function. %
Messages from the neighbourhood $\mathcal{N}(i)$ are then aggregated and combined with the current node features to obtain updated representations,
\begin{equation}\label{eq:aggregate_n_update}
\mathbf{x}_i' = \gamma\!\left(\mathbf{x}_i, \underset{j\in\mathcal{N}(i)}{\square}\mathbf{m}_{ji}; \boldsymbol{\theta}_{\gamma}\right),
\end{equation}
where $\square$ denotes a permutation-invariant aggregation operator and $\gamma(\cdot)$ a learnable update function. %
A schematic illustration of this process is provided in \Cref{fig:message_passing}.

The aggregation operation $\square$ is a \emph{permutation-invariant} function, which allows the message-passing framework to operate consistently on graphs with varying numbers of neighbours. 
Typical choices include element-wise summation, mean, or maximum, each inducing different biases with respect to neighbourhood size and feature scaling.
The choice of the message and update functions $\phi$ and $\gamma$ plays a crucial role in determining the specific architecture and behaviour of a GNN. These choices lead to distinct characteristics that determine their applicability for various physics-problems \cite{sanchez2020learning, de2024multivariate, duthe2025graph}.
Common examples include the \emph{GraphNet} formulation, which employs generic learnable functions for message construction and aggregation and is widely used in physics-based learning tasks \cite{battaglia2018relational,sanchez2020learning,pfaff2020learning}, \emph{Graph Attention Networks} (GATs), which weight messages adaptively using learned attention coefficients \cite{velivckovic2017graph}, and \emph{Graph Isomorphism Networks} (GINs), which use sum aggregation with injective update functions to maximise expressive power \cite{xu2018powerful}.
There are many resources available which detail the characteristics and uses of each of these \cite{zhang2019graph,thomas2022graph}, but for this work, a modified form of the GraphNet, along with a convolutional approach, is used as it is most appropriate for the physics-bias to be introduced, as explained in more detail in \Cref{sec:gnode-struct-dyn}. %

In a general form, the overall function to be learnt can be described as a graph convolutional operator $\mathcal{C}$, which captures the message-aggregation-update process described above,
\begin{equation}
    \mathbf{X}' = \mathcal{C}(\mathcal{G},\boldsymbol{\Theta})
\end{equation}
where $\mathcal{G} = \{\mathbf{X},\mathbf{E},\mathbf{A}\}$ is the graph, described by the full set of node features $\mathbf{X}\in\mathbb{R}^{n_n\times d_{x}}$, full set of edge features $\mathbf{E}\in\mathbb{R}^{n_e\times d_{e}}$, and the adjacency matrix $\mathbf{A}\in\mathbb{N}^{n_n\times n_n}$; a binary valued matrix defining the connections between nodes, as in \cite{poli2019graph}. %
$\boldsymbol{\Theta}=\{\boldsymbol{\theta}_{\phi}, \boldsymbol{\theta}_{\gamma}\}$ are the learnable parameters which govern the chosen message and update functions, $\phi$ and $\gamma$, respectively. %

\subsection{Graph Neural ODE}
\label{sec:gnode}

Following \cite{poli2019graph}, we formulate the graph differential equation (GDE) for a dynamical system in state-space form. Assuming Markovian evolution of the latent states, the continuous-time state transition over a time interval is defined as
\begin{subequations}
    \label{eq:predict_next_state_cont}
    \begin{equation}
        \mathbf{Z}(t + \Delta t) = \mathcal{F}_{\textrm{st}}(\mathbf{Z}(t),\mathbf{F}(t),\mathbf{X}(t),\mathbf{E}(t),\mathbf{A}(t)) = \mathbf{Z}(t) + \int_t^{t+\Delta t} \dot{\mathbf{Z}}(\tau)d\tau
    \end{equation}
    \begin{equation}
        \dot{\mathbf{Z}}(\tau)=\mathcal{F}_{\textrm{ev}}\left(\mathbf{Z}(\tau), \mathbf{X}(\tau),\mathbf{E}(\tau),\mathbf{A}(\tau),\mathcal{F}_{\textrm{in}}(\mathbf{F}(\tau), \mathbf{X}(\tau),\mathbf{E}(\tau),\mathbf{A}(\tau))\right)
    \end{equation}
\end{subequations}
where $\mathbf{Z} \in \mathbb{R}^{n_n\times d_{z}}$ denotes the hidden states of the graph, $\mathbf{F}$ represents any external input, $\mathcal{F}_{\textrm{ev}}$ defines the state evolution (vector field), and $\mathcal{F}_{\textrm{in}}$ denotes the system input mapping. 

In this formulation, $\mathcal{F}_{\textrm{ev}}$ governs the continuous-time evolution of the latent state, while $\mathcal{F}_{\textrm{in}}$ provides a transformation of the external inputs into a representation suitable for the evolution dynamics. The overall state transition function $\mathcal{F}_{\textrm{st}}$ is therefore not an independent model, but is implicitly defined through time integration of the evolution function. Both $\mathcal{F}_{\textrm{ev}}$ and $\mathcal{F}_{\textrm{in}}$ are expressed as functions of the node, edge, and adjacency representations to allow for flexible modelling of graph-structured systems.

Depending on the type of available measurements, an observation (readout) function is introduced to map latent states to measurable quantities,
\begin{equation}
    \mathbf{Y}(t) = \mathcal{F}_{\textrm{o}}(\mathbf{Z}(t),\mathbf{X}(t), \mathbf{E}(t), \mathbf{A}(t))
\end{equation}

In discrete form, the \emph{state transition} function used to estimate the current hidden state from the previous time point is written as
\begin{equation}
\begin{split}
    \hat{\mathbf{Z}}_t &= \mathcal{F}_{\textrm{st}}(\mathbf{Z}_{t-1}, \mathbf{F}_{t-1}, \mathbf{X}_{t-1}, \mathbf{E}_{t-1}, \mathbf{A}_{t-1}) \\
     & = \mathbf{Z}_{t-1} + \int_{t-1}^t \mathcal{F}_{\textrm{ev}}\left(\mathbf{Z}(\tau), \mathbf{X}(\tau),\mathbf{E}(\tau),\mathbf{A}(\tau),\mathcal{F}_{\textrm{in}}(\mathbf{F}_{t-1}, \mathbf{X}_{t-1},\mathbf{E}_{t-1},\mathbf{A}_{t-1})\right) d\tau
\end{split}
\end{equation}

Graph Neural ODEs (GNODEs) aim to \emph{learn} the functions $\mathcal{F}_{\textrm{ev}}$, $\mathcal{F}_{\textrm{in}}$, and $\mathcal{F}_{\textrm{o}}$ using graph neural network architectures. In this context, $\mathbf{V}'$ denotes the encoded input, obtained from the measurable input $\mathbf{F}$ through the input function $\mathcal{F}_{\textrm{in}}$. This mapping may be realised through an input graph convolutional operator $\mathcal{C}_{\textrm{in}}$, derived either from known physics (e.g.\ ground acceleration) or directly from measured system inputs (e.g.\ force). 

The evolution and observation functions are then parameterised using graph convolutional operators $\mathcal{C}_{\textrm{ev}}$ and $\mathcal{C}_{\textrm{o}}$, respectively,
\begin{subequations}
    \begin{equation}
    \mathbf{V}'_t = \hat{\mathcal{F}}_{\textrm{in}}(\mathbf{F}_t, \mathbf{X}_t,\mathbf{E}_t,\mathbf{A}_t) = \mathcal{C}_{\textrm{in}}(\mathbf{F}_t, \mathcal{G}_t; \boldsymbol{\Theta}_{\textrm{in}})
    \end{equation}
    \begin{equation}
    \hat{\mathcal{F}}_{\textrm{ev}}(\tau) = \mathcal{C}_{\textrm{ev}}(\mathbf{Z}(\tau), \mathbf{V}'_t, \mathcal{G}(\tau); \boldsymbol{\Theta}_{\textrm{ev}})
    \end{equation}
    \begin{equation}
    \hat{\mathcal{F}}_{\textrm{o},t} = \mathcal{C}_{\textrm{o}}(\mathbf{Z}_t, \mathcal{G}_t; \boldsymbol{\Theta}_{\textrm{o}})
    \end{equation}
\end{subequations}
where $\mathcal{G} = (\mathbf{X}, \mathbf{E}, \mathbf{A})$ denotes the graph structure and $\boldsymbol{\Theta}$ the learnable parameters.

Importantly, during the integration scheme, for dynamic graphs, the instantaneous graph features are evaluated continuously within the convolution operators, ensuring that the evolution is consistent with the time-varying graph representation over the integration window. 

Given the current estimate of the state, the state at the next time point is computed as
\begin{equation}
    \begin{split}
        \hat{\mathbf{Z}}_t &= \mathcal{F}_{\textrm{st}}(\hat{\mathbf{Z}}_{t-1},\mathbf{X}_{t-1}, \mathbf{F}_{t-1}, \mathbf{E}_{t-1}, \mathbf{A}_{t-1}) \\
         & = \hat{\mathbf{Z}}_{t-1} + \textrm{ODEINT}\left(\hat{\mathcal{F}}_{\textrm{ev}}\left(\mathbf{Z}_t, \mathbf{X}_t,\mathbf{E}_t,\mathbf{A}_t,\mathcal{F}_{\textrm{in}}(\mathbf{F}_{t-1}, \mathbf{X}_{t-1},\mathbf{E}_{t-1},\mathbf{A}_{t-1})\right); \; t-1, t \right)
    \end{split}
\end{equation}
where $\textrm{ODEINT}$ denotes a differentiable numerical integrator, such as Runge–Kutta or Velocity Verlet, the latter being used in this work.

An autoregressive GDE recursively predicts the next state based on previous predictions, similar to recurrent dynamical models. Consequently, accurate prediction depends on a reliable estimate of the initial state. During training, the model parameters are optimised by minimising the discrepancy between predicted observations $\hat{\mathbf{Y}}=\hat{\mathcal{F}}_{\textrm{o}}$ and measured observations $\mathbf{Y}^*$ in the observation domain $\Omega_o$,
\begin{equation}
    L_{\textrm{obs}} = \sum_{s=1}^{n_s} \left|\left|\left(\hat{\mathbf{Y}}[s] - \mathbf{Y}^*[s] \right) \right|\right|^2_{\Omega_o}
    \label{eq:obs_loss}
\end{equation}

\begin{figure}[t!]
    \centering
    \includegraphics[width=0.99\linewidth]{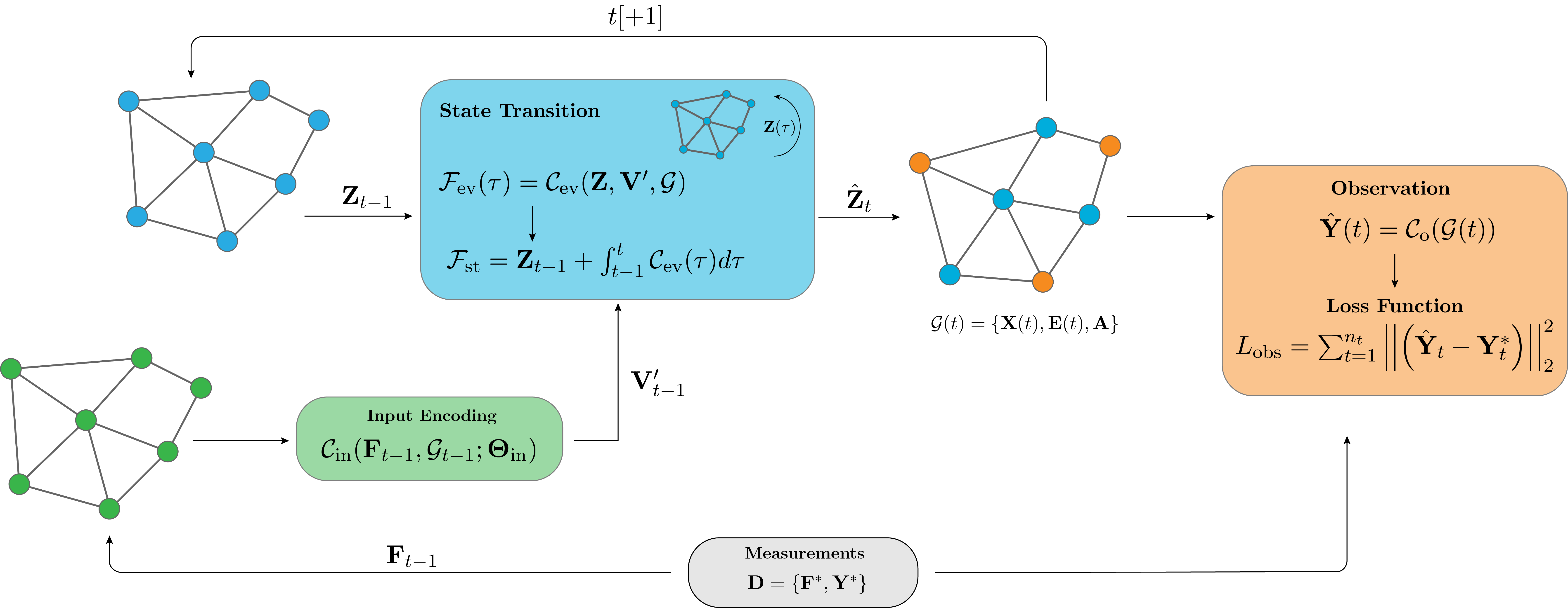}
    \caption{Block diagram of the Graph Neural ODE (GNODE) model, showing the input encoding $\mathcal{F}_{\textrm{in}}$, continuous-time state evolution governed by $\mathcal{F}_{\textrm{ev}}$ on the graph, and the observation mapping $\mathcal{F}_{\textrm{o}}$ used to recover measurable quantities.}
    \label{fig:gnode-model}
\end{figure}

\subsection{GNODEs for Structural Dynamics}
\label{sec:gnode-struct-dyn}

We begin by defining the graph-state-space model for structural dynamics problems in a black-box form, i.e.\ no prior physics knowledge. %
For two-dimensional structural dynamics problems, the hidden state $\mathbf{z}$, the input (external force) $\mathbf{f}$, and observations $\mathbf{y}$ are defined in Cartesian coordinates; for node $i$,
\begin{equation}
    \mathbf{z}_i = \left\{ u_{i,x}, u_{i,y}, \dot{u}_{i,x}, \dot{u}_{i,y} \right\}^T, \quad
    \mathbf{f}_i = \{f_{i,x}, f_{i,y}\}^T, \quad
    \mathbf{y}_i = \{u_{i,x}, u_{i,y}, \dot{u}_{i,x}, \dot{u}_{i,y}, \ddot{u}_{i,x}, \ddot{u}_{i,y}\}^{T}
\end{equation}
where $u_d$, $\dot{u}_d$, $\ddot{u}_d$, and $f_d$ are the nodal displacement, velocity and acceleration and the external force in the direction $d$, respectively. %
The general \ac{gnode} formulation allows separate learnable input and state-evolution functions, providing maximum flexibility. %
However, for structural dynamics problems this formulation can be simplified by embedding the system states and external inputs directly within the graph representation. Specifically, we assume that the system states and external inputs are embedded within, or deterministically mapped to, the node and edge feature spaces, i.e.\ $\mathbf{Z}, \mathbf{F} \subseteq \{\mathbf{X}, \mathbf{E}\}$, or more generally that certain graph features are functions of the states and/or external inputs, $f(\mathbf{Z}, \mathbf{F}) \subseteq \{\mathbf{X}, \mathbf{E}\}$. Here, $\mathbf{Z}$ and $\mathbf{F}$ denote the flattened state and force vectors over the full graph.

For structural dynamics, the adjacency matrix is assumed time-invariant, and the state-evolution and observation equations reduce to,
\begin{subequations}
\begin{equation}
    \dot{\mathbf{Z}}(\tau)=\mathcal{F}_{\textrm{ev}}(\mathbf{X}(\tau),\mathbf{E}(\tau),\mathbf{A})
\end{equation}
\begin{equation}
    \mathbf{Y}(\tau)=\mathcal{F}_{\textrm{o}}(\mathbf{X}(\tau),\mathbf{E}(\tau),\mathbf{A})
\end{equation}
\end{subequations}
The state transition function $\mathcal{F}_{\textrm{st}}$ is not learned independently, but is implicitly defined through time integration of the graph-based state evolution function $\mathcal{F}_{\textrm{ev}}$, as introduced in \Cref{sec:gnode}.

A distinct advantage of \acp{gnn} is their inherent inductive biases which stem from their topology-aware architecture. %
Additional inductive biases can also be introduced in the form of physics-informed loss functions \cite{ngo2024physics}, and for dynamical systems, one can manipulate the architecture to embed further inductive biases via representations of force balances \cite{bishnoi2022enhancing, thangamuthu2022unravelling}. %
However, for nonlinear structural dynamics, and other problems where the underlying physics may be partially known, there is an opportunity to exploit the advantages of \emph{physics-guided machine learning} (PGML). %
In PGML schemes, there is a trade-off between a lower reliance on data for a larger reliance on the underlying physics model. %

The first step for designing biased GNODEs for structural dynamics is the selection of appropriate features throughout the graph. %
An important consideration here, is to ensure \emph{equivariance} of the graph model, such that it is not affected by transformations of the reference coordinate system \cite{Batzner_2022,satorras2022enequivariantgraphneural,sharma2025equi}. %
First, the node features are set as the rest position $\mathbf{p}_0$, external force $\mathbf{f}$ and equivalent mass $m$, and the edge features as the connection extensions $\epsilon$, connection extension rates $\dot{\epsilon}$, cosine and sine of the current bar angles $\theta$, cosine and sine of the rest angles $\phi_{ji}$, stiffness $k$, and damping $c$. %
The cosine and sine are used in order to remove discontinuities in the input features around $\pm\pi/2$ and represent normalised Euclidean directions of the edge. %
\begin{equation}
    \mathbf{x}_i = \{ p_{0,x,i}, p_{0,y,i}, f_{x,i}, f_{y,i}, m_i \} \qquad \mathbf{e}_{ji} = \{ \epsilon_{ji}, \dot{\epsilon}_{ji}, \cos(\theta_{ji}), \sin(\theta_{ji}), \cos(\phi_{ji}), \sin(\phi_{ji}), k_{ji}, c_{ji} \}
\end{equation}
where the values of $\epsilon_{ji}, \dot{\epsilon}_{ji}, \theta_{ji}$ are continuously updated using corotational kinematic calculations on the current node states $\mathbf{z}_i=[\mathbf{u}_i,\dot{\mathbf{u}}_i]^T$, further details of which are given in Appendix A. %
In the context of structural mechanics, these geometric nonlinearities are simple to include and a powerful inductive bias, however, one could also omit this prescription, and allow the learner freedom in capturing more complex/unknown geometric nonlinearities by passing the states as part of the node features. %

The state evolution function (as a graph convolution operator) is augmented to be a function $\mathcal{F}$ of the physics-based (linear) graph convolution $\mathcal{C}_{\textrm{phy}}(\mathcal{G})$, estimated from the known structural mechanics, and black-box neural-network-based graph convolution $\mathcal{C}_{\textrm{bb}}(\mathcal{G},\boldsymbol{\Theta})$, estimated by the \ac{gnn} forward pass,
\begin{equation}
    \mathcal{F}_{\textrm{ev}}(\mathbf{Z}, \mathbf{F}, \mathcal{G}, \boldsymbol{\Theta}) = \mathcal{F}(\mathcal{C}_{\textrm{phy}}(\mathbf{Z}, \mathbf{F}, \mathcal{G}), \mathcal{C}_{\textrm{bb}}(\mathbf{Z}, \mathbf{F}, \mathcal{G},\boldsymbol{\Theta}))
\end{equation}
The important choice here is that of the graph convolutional operator $\mathcal{C}_{\textrm{bb}}$ (i.e. the message-aggregation-update scheme), where further biases can be introduced by architecture design. %

\begin{figure}
    \centering
    \includegraphics[width=0.99\linewidth]{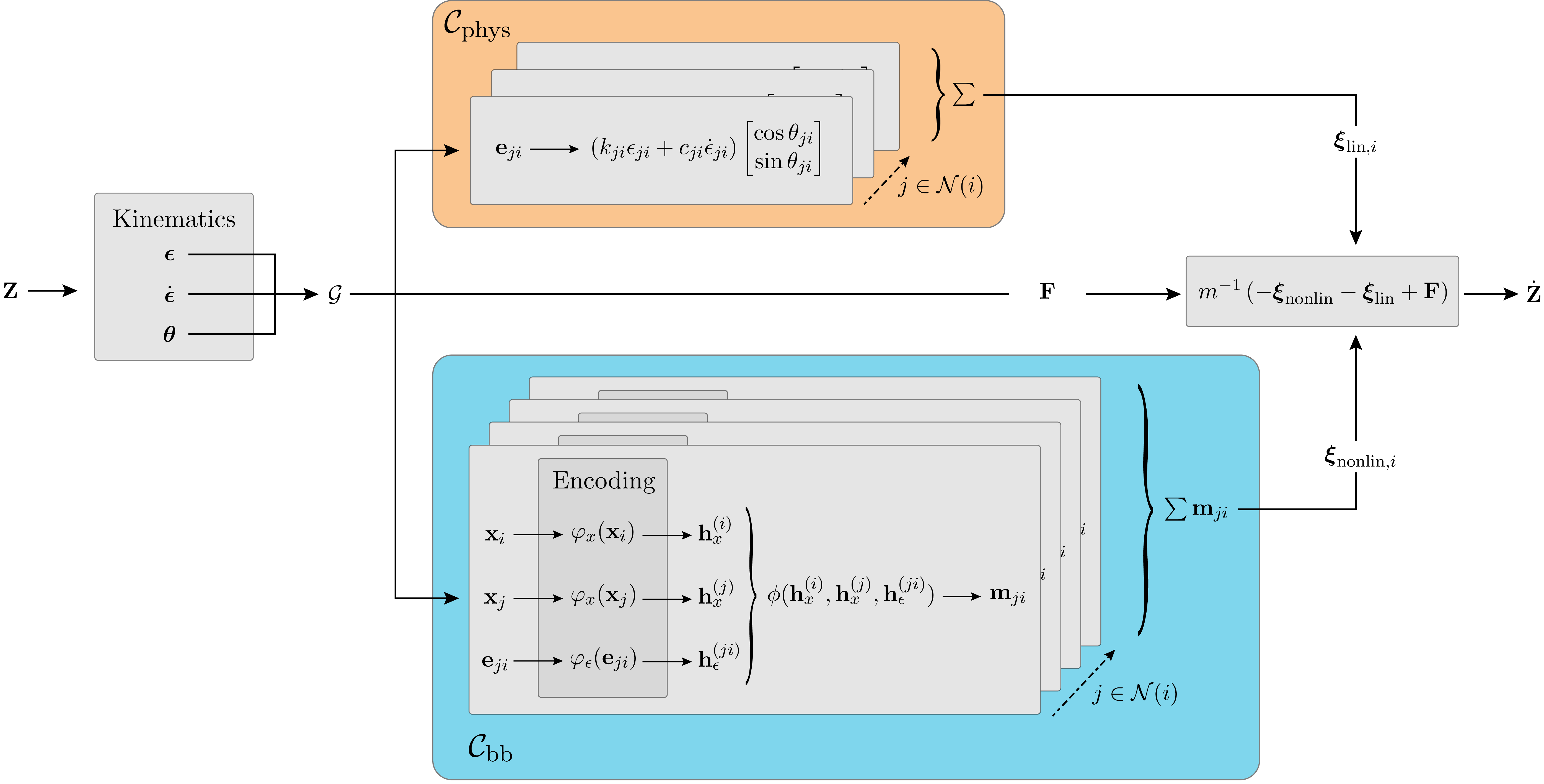}
    \caption{Block diagram of the \ac{pggnode}, showing the estimation of the state evolution function $\mathcal{F}_{\textrm{ev}}$ (\Cref{eq:state_deriv}) via the combination of physics-based ($\mathcal{C}_{\textrm{phy}}$) and data-driven ($\mathcal{C}_{\textrm{bb}}$) graph convolutional operators.}
    \label{fig:piggo-blockdiag}
\end{figure}

One such bias permissible, which is used in this work, is to assume the nonlinearity manifests as a conservative force acting on node $i$ by edge $ji$, thus the message aggregation is a sum. %
In this work, the node and edge features ($\mathbf{x}$ and $\mathbf{e}$) are encoded into their respective latent variables $\mathbf{h}_x$ and $\mathbf{h}_{\epsilon}$ before passing in to the message function, in order to increase flexibility of the learnt function. %
\begin{subequations}
\begin{equation}
    \mathbf{h}_{x}^{(i)} = \varphi_x(\mathbf{x}_i) = \mathrm{LeakyReLU}(\mathbf{x}_i), \quad \mathbf{h}^{(ji)}_{\epsilon} = \varphi_{\epsilon}(\mathbf{e}_{ji}) = \mathrm{LeakyReLU}(\mathbf{e}_{ji})
\end{equation}
\begin{equation}
    \mathbf{m}_{ji} = \phi_m(\mathbf{x}_j, \mathbf{x}_i, \mathbf{e}_{ji}) = \mathrm{LeakyReLU} (\{\mathbf{h}_{x}^{(j)}, \mathbf{h}_{x}^{(i)}, \mathbf{h}_{\epsilon}^{(ji)}\})
\end{equation}
\begin{equation}
    \boldsymbol{\xi}_{\textrm{nonlin},i} = \mathcal{C}_{\textrm{bb}}(\mathcal{G}) = \underset{j\in\mathcal{N}(i)}{\square}\hat{\mathbf{n}}_{ji}\mathbf{m}_{ji} = \sum_{j\in\mathcal{N}_i}\hat{\mathbf{n}}_{ji}\phi_m(\mathbf{x}_j, \mathbf{x}_i, \mathbf{e}_{ji})
\end{equation}
\end{subequations}
where $\boldsymbol{\xi}_{\textrm{nonlin},i}$ represents the nonlinear restoring forces, and $\hat{\mathbf{n}}_{ji} = [\cos(\theta_{ji}),\sin(\theta_{ji})]^T$ is the current direction of the edge. %
The message-passing MLP is chosen to use the LeakyReLU activation function to avoid singularities/gradient explosion, as the values of the kinematics often fluctuate around zero. %
For the full system state, the evolution is then an additive combination of the known linear and external forces, and unknown nonlinear forces,
\begin{equation}
    \dot{\mathbf{z}}_i = \begin{bmatrix}\dot{\mathbf{u}}_i \\
    m_i^{-1}\left(-\mathcal{C}_{\textrm{phy}}^{(i)}(\mathbf{Z},\mathcal{G}) - \mathcal{C}^{(i)}_{\textrm{bb}}(\mathbf{Z}, \mathcal{G}, \boldsymbol{\Theta})\right)
    \end{bmatrix} + \begin{bmatrix}
        \boldsymbol{0} \\ \mathbf{f}_i
    \end{bmatrix}
    \label{eq:state_deriv}
\end{equation}
For dynamics of point masses, the physics-based (linear) graph convolutional operator can be formulated from the force balance at each node, %
\begin{equation}
    \mathcal{C}_{\textrm{phy}}^{(i)}(\mathbf{Z},\mathcal{G}) = \boldsymbol{\xi}_{\textrm{lin},i} = \sum_{j\in\mathcal{N}(i)} (k_{ji}\epsilon_{ji} + c_{ji}\dot{\epsilon}_{ji})\begin{bmatrix}
        \cos\theta_{ji} \\
        \sin\theta_{ji} \\
    \end{bmatrix}
\end{equation}
where $\boldsymbol{\xi}_{\textrm{lin},i}$ is the sum of the restoring forces exerted on the point mass by all connected edges. %
This formulation can be directly compared to the aggregation in \Cref{eq:aggregate_n_update}, where the operation can be interpreted as a physics-based message aggregation of restoring forces.
The process block diagram showing the \ac{pggnode} architecture is shown in \Cref{fig:piggo-blockdiag}, which represents the total convolutional approach to continuously estimate the state evolution, $\dot{\mathbf{Z}}$, given the current estimate of the state, $\mathbf{Z}$, (\Cref{eq:state_deriv}). %

At this stage, the observation-based loss function defined in \Cref{eq:obs_loss} could be employed. However, to enable optimisation within a physics-consistent objective space, we instead reformulate it as a physics-informed loss function.
To do so, given the predicted states and observed values of the acceleration and force at time point $t$, the physics loss then minimises a force-balance residual
\begin{equation}
    \mathbf{r}_i(\tilde{\mathbf{Z}}(t), \mathbf{f}_i^*(t), \ddot{\mathbf{u}}_i^*(t)) = m_i\ddot{\mathbf{u}}_i^*(t) + \boldsymbol{\xi}_i(\tilde{\mathbf{Z}}(t)) - \mathbf{f}_i^*(t)
    \label{eq:phys_residual}
\end{equation}
\begin{equation}
    L_{\textrm{phy}} = \sum_{s=1}^{n_t} \left|\left| \mathbf{R}(\tilde{\mathbf{Z}}(t), \mathbf{F}^*(t), \ddot{\mathbf{U}}^*(t)) \right|\right|^2 = \sum_{t=1}^{n_t}\sum_{i=1}^{n_n} \left|\left| m_i\ddot{\mathbf{u}}_i^* + \boldsymbol{\xi}_i - \mathbf{f}_i^* \right|\right|^2
    \label{eq:phys_det_loss}
\end{equation}
where $\boldsymbol{\xi}_i(\tilde{\mathbf{Z}}(t)) = \boldsymbol{\xi}_{\textrm{lin},i}(\tilde{\mathbf{Z}}(t)) + \boldsymbol{\xi}_{\textrm{nonlin},i}(\tilde{\mathbf{Z}}(t))$ is the sum of the linear and nonlinear restoring forces, based on the full graph state $\tilde{\mathbf{Z}}(t)$, as calculated in the graph convolutional process. %
The architecture provides two primary sources of physics-based inductive biases; the topologically-aware and locally-coherent operations, and the physics-guided state evolution prediction. %
The former enables learning and prediction from spatially sparse observations, whereas the latter introduces a physics-consistent objective that is partly redundant with the observation loss.
To remove this redundancy, one may wish to use the predicted accelerations rather than the observed accelerations in the physics loss, however, this would create a tautology as the predicted accelerations come from the same underlying physics equations. %
Therefore, in this work, only the physics loss is used in all training, unless otherwise stated. %
\begin{equation}
    L = L_{\textrm{phy}}
\end{equation}
Compared to traditional physics-informed approaches, this makes for simpler training as it removes any requirement for manual tuning of loss weights. %


\begin{algorithm}
    \caption{Physics-guided GNODE algorithm for virtual sensing with acceleration observation model.}
    \label{alg:pggnode}
    \hspace*{\algorithmicindent} \textbf{Input} Measured Data $\mathbf{D}^*$, Structural Information $\mathcal{D}$ \\
    \hspace*{\algorithmicindent} \textbf{Output} Predictions of State Over Entire Graph Domain
    \begin{algorithmic}[1] 
        \State $\mathbf{t}^*, \ddot{\mathbf{U}}^*, \mathbf{F}^* \gets \mathbf{D}^*$
        \State $\mathbf{X},\mathbf{E},\mathbf{A} \gets \mathcal{D}$
        \State Initialise GNODE $\mathcal{G}(\mathbf{X},\mathbf{E},\mathbf{A};\boldsymbol{\Theta})$
        \State Initialise network weight optimiser \texttt{optim\_net}$(\boldsymbol{\Theta})$
        \For{$i = 0$ to epochs}
            \State $\hat{\mathbf{Z}} = \emptyset^{n_t\times n_n \times n_z}$  \Comment{Initialise tensor to store state predictions}
            \State $\hat{\mathbf{Y}} = \emptyset^{n_t\times n_n \times n_y}$ \Comment{Initialise tensor to store observations}
            \State $L_{\textrm{phy}} = \emptyset^{n_t}$
            \For{$s=1$ to $n_t$}
                \State $\hat{\mathbf{Z}}[s] \gets \mathbf{Z}[s-1] + \textrm{ODEINT}(\mathcal{C}_{\textrm{ev}};t[s-1],t[s]) \quad|\quad \Omega_p$ \Comment{Estimate state transition}
                \State $L_{\textrm{phy}}[s] \gets ||\mathbf{R}(\hat{\mathbf{Z}}[s], \mathbf{F}^*[s], \mathbf{Y}^*[s])||_{\Omega_o}^2$ \Comment{Calculate physics residual in observation domain}
            \EndFor
            \State $L \gets \frac{1}{n_t} \sum L_{\textrm{phy}}$ \Comment{See loss function in \Cref{eq:phys_det_loss}}
            \State Update $\boldsymbol{\Theta}$ using $\nabla L$ and \texttt{optim\_net}
            \If{$\nabla L \leq \epsilon_L$}
                \State \textbf{end for} \Comment{Early stopping}
            \EndIf
        \EndFor
    \end{algorithmic}
\end{algorithm}

\section{Graph-based Bayesian Virtual Sensing under Sparse and Noisy Data}

Section 2 presented the architecture through which the dynamics of systems with partially known and locally consistent physical behaviour can be learned. Owing to the inductive biases introduced, the resulting model can be directly employed in an offline setting. In this section, we extend this framework by integrating it with a Kalman filtering approach to enable online state estimation.

\subsection{Graph Kalman Filters}

In civil and mechanical engineering, a popular choice for virtual sensing of weakly nonlinear systems is the Extended Kalman Filter \cite{chui2009kalman}, because of its ease of use, robustness and suitability for real-time applications. %
The process assumes a sequence of available measurements $\mathbf{y}_{1:T}$, which are determined by some hidden states $\mathbf{z}_{1:T}$. %
The transition from a state $\mathbf{z}_{t-1}$ to the next state $\mathbf{z}_t$ is denoted the \emph{transition} function, and the process from a state $\mathbf{z}_t$ to its observable variable $\mathbf{y}_t$ is termed the \emph{observation} function. %
A full introduction and tutorial for the \ac{ekf}, and other KF variants, can be found in \cite{kim2018introduction}, but an important consideration is that it assumes both these functions as known,
\begin{subequations}
    \begin{equation}
        \mathbf{z}_t = f_{\textrm{st}}(\mathbf{z}_{t-1}, \mathbf{f}_{t-1}) + w_t
    \end{equation}
    \begin{equation}
        \mathbf{y}_t = f_{\textrm{o}}(\mathbf{z}_t) + v_t
    \end{equation}
\end{subequations}
where $w\sim\mathcal{N}(0,\mathbf{Q}_t)$ and $v\sim\mathcal{N}(0,\mathbf{R}_t)$ are Gaussian noise sources for the process and observation, respectively. %

Kalman filtering schemes are reliant on the underlying model which defines the state transition $f_{\textrm{st}}$ and observation $f_{\textrm{o}}$ functions. %
Following the graph differential equation formulation in \Cref{sec:gnode}, these functions are parameterised through graph-based operators, where the state transition $\mathcal{F}_{\textrm{st}}$ is induced via integration of the evolution function $\mathcal{F}_{\textrm{ev}}$, and the observation function $\mathcal{F}_{\textrm{o}}$ is defined through a graph readout operator. %
In the graph setting, these quantities are lifted to graph-level representations, where $\mathbf{Z}_t$ and $\mathbf{Y}_t$ denote the stacked states and observations over all nodes. Accordingly, the filtering equations become,
\begin{subequations}
    \begin{equation}
        \mathbf{Z}_t = \mathcal{F}_{\textrm{st}}(\mathbf{Z}_{t-1},\mathbf{F}_{t-1},\mathcal{G}_{t-1}) + \boldsymbol{W}_t 
    \end{equation}
    \begin{equation}
        \mathbf{Y}_t = \mathcal{F}_{\textrm{o}}(\mathbf{Z}_t,\mathcal{G}_t) + \boldsymbol{V}_t 
    \end{equation}
\end{subequations}
where $\boldsymbol{W}_t$ and $\boldsymbol{V}_t$ are the process and measurement noise terms for all nodes on the graph. %
The prior distribution of the states over the graph is given by,
\begin{subequations}
    \begin{equation}
        q(\mathbf{Z}_t|\mathbf{Y}_{1:t-1},\mathbf{F}_{1:t-1}) = \mathcal{N}(\boldsymbol{\mu}_{t|t-1},\boldsymbol{\Sigma}_{t|t-1})
    \end{equation}
    \begin{equation}
        \boldsymbol{\mu}_{t|t-1} = \mathcal{F}_{\textrm{st}}(\boldsymbol{\mu}_{t-1|t-1},\mathbf{F}_{t-1},\mathcal{G}_{t-1|t-1})
        \label{eq:prior_mean_gkf}
    \end{equation}
    \begin{equation}
        \boldsymbol{\Sigma}_{t|t-1} = \mathbf{A}_{t-1}\boldsymbol{\Sigma}_{t-1|t-1}\mathbf{A}_{t-1}^T + \mathbf{Q}_{t-1}
        \label{eq:prior_covar_gkf}
    \end{equation}
\end{subequations}
where $\mathbf{A}_{t-1} = \nabla_{\mathbf{Z}} \mathcal{F}_{\textrm{st}}(\mathbf{Z}_{t-1}, \mathbf{F}_{t-1}, \mathcal{G}_{t-1})$ is the Jacobian of the state transition function at the previous posterior estimate. %
Then, the posterior distribution for the \emph{update} step is computed,
\begin{subequations}
    \begin{equation}
        \mathbf{K}_t = \boldsymbol{\Sigma}_{t|t-1}\mathbf{H}_t^T\left( \mathbf{H}_t\boldsymbol{\Sigma}_{t|t-1}\mathbf{H}_t^T + \mathbf{R}_t \right)^{-1}
        \label{eq:kgain_gkf}
    \end{equation}
    \begin{equation}
        \boldsymbol{\mu}_{t|t} = \boldsymbol{\mu}_{t|t-1} + \mathbf{K}_t \left[ \mathbf{Y}_t^* - \mathcal{F}_{\textrm{o}}(\boldsymbol{\mu}_{t|t-1},\mathcal{G}_{t|t-1}) \right]
        \label{eq:post_mean_gkf}
    \end{equation}
    \begin{equation}
        \Sigma_{t|t} = (\mathbf{I} - \mathbf{K}_t\mathbf{H}_t)\Sigma_{t|t-1}
        \label{eq:post_covar_gkf}
    \end{equation}
\end{subequations}
where $\mathbf{H}_t = \nabla_{\mathbf{Z}} \mathcal{F}_{\textrm{o}}(\mathbf{Z}_t, \mathcal{G}_t)$ is the Jacobian of the observation function at the current prior estimate. %
This posterior is then inserted into the graph feature location to create the posterior estimate of the graph $\mathcal{G}_{t|t}$. %
In practice, with modern machine learning libraries, such as Pytorch or JAX, the Jacobians can be returned easily using automatic differentiation. %
As can be seen in the above equations, the Kalman filtering procedure is performed in a recurrent/autoregressive manner, where previous predictions are used directly in the prediction of the current step. %
A powerful advantage of KF schemes is their robustness against unknown initial states, as the continuous updating scheme `settles' to an accurate solution over time. %

\begin{figure}
    \centering
    \includegraphics[width=0.99\linewidth]{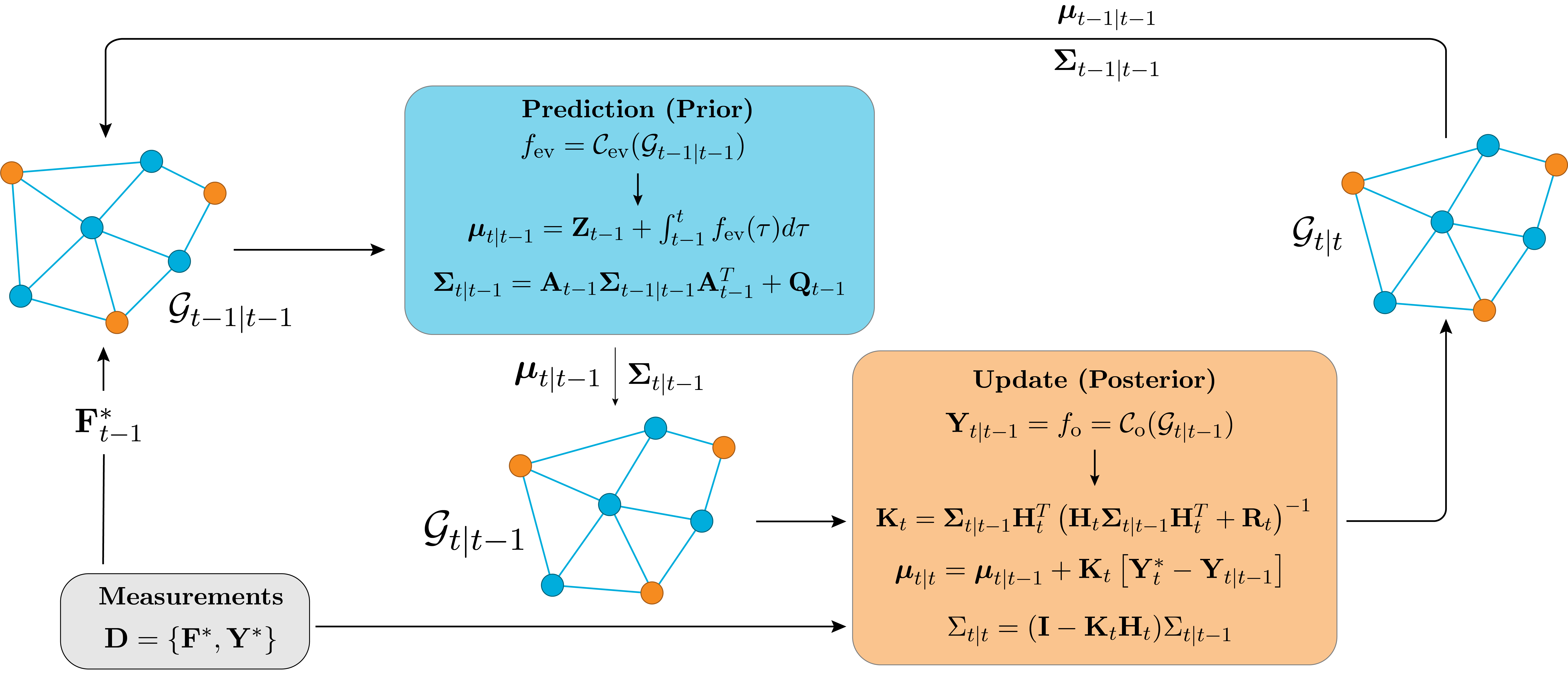}
    \caption{Block diagram of the graph extended Kalman filter.}
    \label{fig:gekf-blockdiag}
\end{figure}

\begin{algorithm}
    \caption{Graph extended Kalman filter algorithm for virtual sensing of structure.}
    \label{alg:gkf}
    \hspace*{\algorithmicindent} \textbf{Input} Measured Data $\mathbf{D}^*$, Model Information $\mathcal{D}$ \\
    \hspace*{\algorithmicindent} \textbf{Output} Predictions of State Over Entire Graph Domain
    \begin{algorithmic}[1] 
        \State $\mathbf{t}^*, \mathbf{Y}^*, \mathbf{F}^* \gets \mathbf{D}^*$
        \State $\mathbf{X},\mathbf{E},\mathbf{A} \gets \mathcal{D}$
        \State Initialise graph model $\mathcal{G}(\mathbf{X},\mathbf{E},\mathbf{A};\boldsymbol{\Theta})$
        \State $\hat{\mathbf{Z}} = \emptyset^{n_s\times n_n \times n_z}$  \Comment{Initialise tensor to store state predictions}
        \State $\hat{\mathbf{Y}} = \emptyset^{n_s\times n_n \times n_y}$  \Comment{Initialise tensor to store state predictions}
        \State $\boldsymbol{\mu}_{0|0} = \emptyset^{n_n \times n_z}$  \Comment{Initialise states of graph to zero}
        \For{$s=1$ to $n_s$}
            \State $t = \mathbf{t}[s]$
            \State Predict prior distribution $\boldsymbol{\mu}_{t|t-1}, \Sigma_{t|t-1}$ using \Cref{eq:prior_mean_gkf,eq:prior_covar_gkf}
            \State Compute Kalman gain $K_t$ using \Cref{eq:kgain_gkf}
            \State Predict posterior distribution $\boldsymbol{\mu}_{t|t}, \Sigma_{t|t}$ using \Cref{eq:post_mean_gkf,eq:post_covar_gkf}
            \State Update graph if necessary $\mathcal{G}_{t|t} \gets \boldsymbol{\mu}_{t|t}, f(\boldsymbol{\mu}_{t|t})$
            \State Store posterior predictions $\hat{\mathbf{Z}}[s] \gets \boldsymbol{\mu}_{t|t}, \; \hat{\mathbf{Y}}[s] \gets \mathcal{F}_{\textrm{o}}(\boldsymbol{\mu}_{t|t}, \mathcal{G}_{t|t})$
        \EndFor
    \end{algorithmic}
\end{algorithm}

\subsection{Learnable Graph Kalman Filters}

As the Graph Kalman Filter is inherently probabilistic, the deterministic loss functions introduced in \Cref{sec:off-virt-sens} do not explicitly account for uncertainty.
Therefore, the first task in creating the learnable \acp{gkf} is to adapt the loss function to take into account noise and model error. %
In classical machine learning, likelihood-based loss functions are often employed to mitigate overfitting; however, in the present context, their primary role is to explicitly account for uncertainty in the residual dynamics. %
First, the physics-likelihood can be formulated using the force balance residual, which is assumed to have a zero-mean Gaussian distribution $p(\mathbf{r}) = \mathcal{N}(\mathbf{0}, \boldsymbol{\Sigma}_r)$, where the residual covariance is defined as
\begin{equation}
    \boldsymbol{\Sigma}_r = \mathrm{diag}(\sigma_{r,1}^2, \dots, \sigma_{r,n_n}^2)
\end{equation}
and $\sigma_{r,i}^2$ denotes the variance of the residual at node $i$, capturing the effects of measurement noise and model error.

Given the residual calculation in \Cref{eq:phys_residual}, and assuming conditional independence of the residuals across nodes, the likelihood can be written as
\begin{equation}
p(\mathbf{F}^*_t,\ddot{\mathbf{U}}^*_t \mid \tilde{\mathbf{Z}}_t;\boldsymbol{\Theta})
= \prod_{i=1}^{n_n} \frac{1}{\sqrt{2\pi \sigma_{r,i}^2}}
\exp \left(-\frac{1}{2}
\frac{\left\| \mathbf{r}_i(\tilde{\mathbf{Z}}_t, \mathbf{F}^*_t, \ddot{\mathbf{U}}^*_t) \right\|^2
}{\sigma_{r,i}^2}
\right)
\end{equation}
where the estimated state one step ahead is given by the prediction process over the graph in \Cref{eq:predict_next_state_cont}. %
During training, the residual variance is estimated from the measurement noise of the acceleration and force,
\begin{equation}
    \sigma_{r,i}^2 = m_i^2 \sigma_{\ddot{u},i}^2 + \sigma_{f,i}^2
    \label{eq:residual_var}
\end{equation}
This derivation assumes that the residual variance is fully attributable to measurement noise, thereby neglecting explicit modelling of model-form uncertainty.
The physics-loss is given using the negative-log-likelihood of the observations based on the predicted force balance at the current time step,
\begin{equation}
    L_{\textrm{phy}} = \sum_{t=1}^{n_t} -\log p(\mathbf{F}^*_t,\ddot{\mathbf{U}}^*_t \mid \tilde{\mathbf{Z}}_t;\boldsymbol{\Theta})
    \label{eq:likeli_loss}
\end{equation}

The acceleration and force variances can be included in the set of learnable parameters in order to be learned. %
In this work, we assume these observation noise variances can be estimated \emph{a priori}, which can be done practically with measurements at known rest. %

For the work shown in this paper, the initial model training is still performed in the standard physics-guided GNODE workflow outlined in \Cref{alg:pggnode}, followed by the trained model being used in the GEKF workflow as in \Cref{alg:gkf}. %
The reason for this is that the additional computational complexity of an \emph{all-in-one graph-neural-EKF} would greatly increase training cost and challenges. %
One benefit of \emph{all-in-one neural-EKF} for non-topological architectures (such as a standard \ac{rnn}) is that they allow for KF-based virtual sensing paradigms to be included within learning architectures to improve sparse estimates without an inductive bias. %
Here, as there are strong inductive biases, offline virtual sensing through scenario-specific training results in a good estimate of the underlying model for the filtering procedures. %
The purpose of the filtering on this pretrained model, is to improve generalisation over domains which are topologically and functionally similar, but with natural variations which increase the model error. %

\section{Nonlinear Structural Dynamics Case Studies}


To analyse the architecture outlined here, a number of nonlinear structural dynamics problems are tested. %
The first two problems are simulated examples of nonlinear dynamics; the first of which is a random spring-mass array with cubic nonlinearities, and the second of which is a truss-bridge array with an angular-gap nonlinearity. %
For all system setups, observation sparsity was introduced by setting every $n_m$-th node as unmeasured, where $n_m$ is defined by a sparsity percentage $p_s$,
\begin{equation}
    n_m = n_n \times \textrm{round}\left( \frac{p_s}{100} \right)
\end{equation}


\subsection{Sobol array with cubic stiffness}

\begin{figure}[h!]
    \centering
    \includegraphics[width=0.85\linewidth]{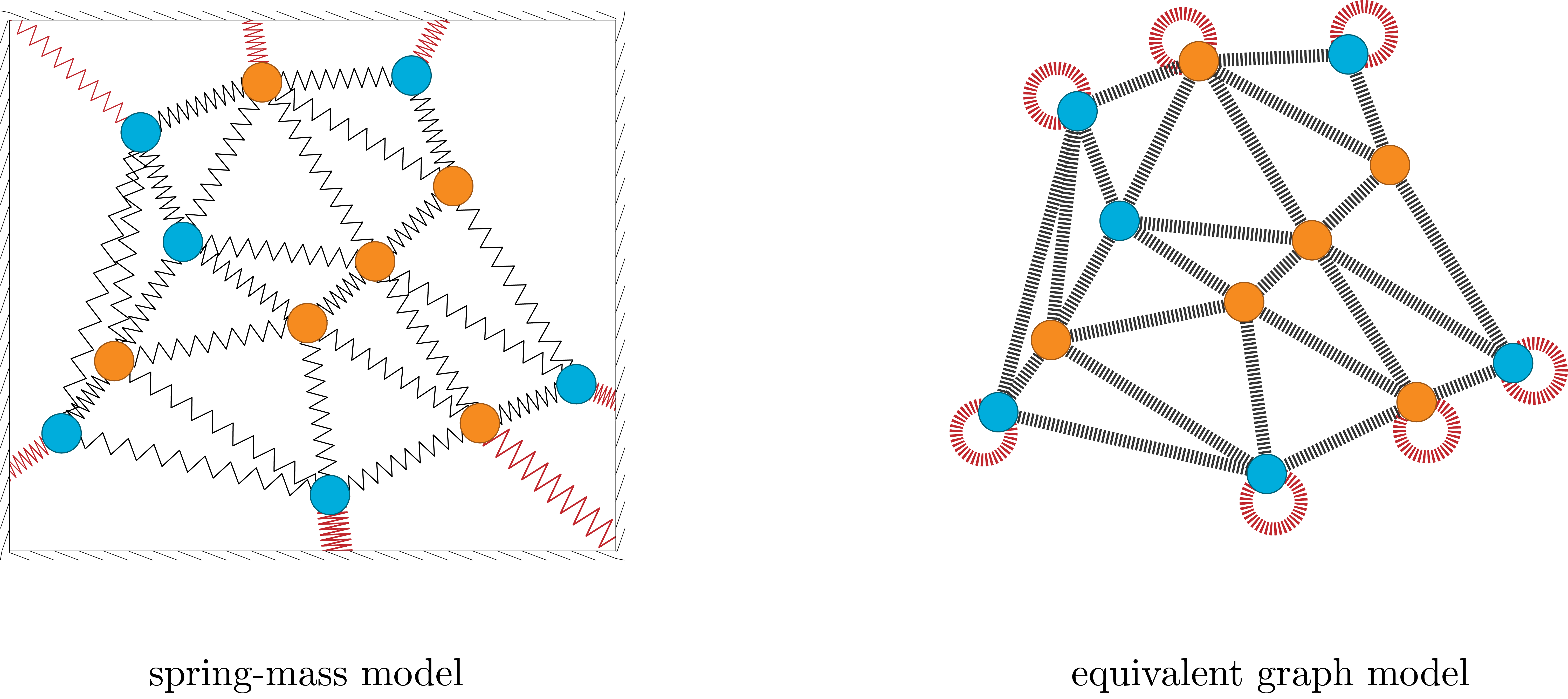}
    \caption{Example of the random truss array with (Left) the spring mass model, where the boundary connections are shown in red and (Right) the equivalent graph model, where the boundary conditions are shown with red self-loops. The orange nodes represent those which are unmeasured.}
    \label{fig:rand-truss-examp}
\end{figure}

The first problem is a random truss array, defined by setting $n_n$ nodes within a 5m x 5m square domain, generated using a Sobol sequence over the two-dimensional plane, which are then connected via Delaunay triangulation. %
The system then treats the connections between node as a spring-damper; for a connection between nodes $i$ and $j$, the spring-damper $ij$ exerts a linear restoring force on node $i$, a nonlinear cubic stiffness restoring force
\begin{equation}
    r_{ij} = k_{ij}\epsilon_{ij} + c\dot{\epsilon}_{ij} + \kappa_{ij}\epsilon^3
\end{equation}
where $k_{ij}$, $c_{ij}$ and $\kappa_{ij}$ are the equivalent spring stiffness, viscous damping, and nonlinear stiffness in the direction of the connection and $r_{ij}$ denotes the scalar restoring force along edge $ij$, which contributes to the nodal force $\boldsymbol{\xi}_i$ in the global coordinate system.
Boundary conditions are then assigned to the outermost (furthest from topological centre) nodes, which are connected with additional spring-damper connections to a fixed boundary of a square, for which the size depends on the number of nodes. %
This system is then simulated with a force input of banded white noise, applied to the lowest 4 nodes, i.e. $f_i(t)=$\textrm{GWN}(0.5rad, 4rad, 2.0s). %
An example of such an array for 12 nodes and a measurement sparsity $p_s=50\%$ is shown in \Cref{fig:rand-truss-examp}. %

\subsection{Bridge truss with angular expansion joints}
The second problem used in this work is a bridge-truss array, which is characterized by non-smooth nonlinearity in the form of an angular clearance stiffness. %
In this setup, the restoring forces acting on node $i$ from truss $ij$ is dependent on the truss angles relative to their rest angle $\phi_{ij}$ and clearance angle $\Phi_{ij}$,
\begin{equation}
    r_{ij} = k_{ij}\epsilon_{ij} + c\dot{\epsilon}_{ij} + 
    \begin{cases}
        0 & \text{if } |\theta_{ij} - \phi_{ij}| < \Phi_{ij} \\
        k_{r,ij}(\theta_{ij} - \phi_{ij}) & \text{if } |\theta_{ij} - \phi_{ij}| \geq \Phi_{ij} \\
    \end{cases}
\end{equation}

\begin{figure}[h!]
    \centering
    \includegraphics[width=0.99\linewidth]{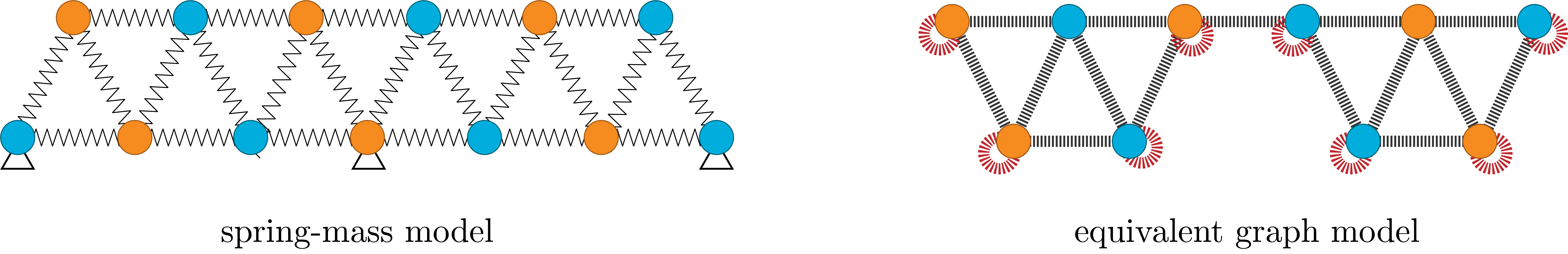}
    \caption{Example of the bridge truss array with (Left) the spring mass model, where the boundary connections are shown in red and (Right) the equivalent graph model, where the boundary conditions are shown with red self-loops. The orange nodes represent those which are unmeasured.}
    \label{fig:bridge-truss-examp}
\end{figure}

The boundary conditions were set as fixed at the outer and centres nodes, on the lowest level -- representing a span bridge with a support column in the centre of the span. %
The equivalent spring-model and its corresponding graph is shown in \Cref{fig:bridge-truss-examp}, where the boundary conditions are embedded in the graph by setting the self-loops on all nodes connected to the fixed nodes of the bridge truss model. %

\subsection{Introducing Uncertainty}

The first and most important source of error in the model is introduced in the form of model uncertainty. %
For both simulation problems, the true structural parameters used in the simulation are sampled from a normal distribution, and the mean values are the nominal values prescribed in the underlying GNODE model,
\begin{equation}
    k_{ij} \sim \mathcal{N}(\mu_k,\sigma_k), \quad c_{ij} \sim \mathcal{N}(\mu_c,\sigma_c)
\end{equation}
The same is done for the nonlinear system parameters, however, these are assumed unknown in the PG-GNODE model,
\begin{equation}
    \kappa_{ij} \sim \mathcal{N}(\mu_{\kappa},\sigma_{\kappa}), \quad k_{r,ij} \sim \mathcal{N}(\mu_{k,r},\sigma_{k,r}), \quad \Phi_{ij} \sim \mathcal{N}(\mu_{\Phi},\sigma_{\Phi})
\end{equation}
Uncertainty is then introduced into the model by setting the standard deviation of the system parameters to 5\% of the mean value, representing a 95\% tolerance interval of $\pm10\%$, the values of which are shown in \Cref{tab:sim-prescr-values}. %
As the edges in the underlying \ac{gnn} act the same function to estimate the nonlinear functions, which in truth have varying parameters but the same model form, they effectively perform mean field approximation of the nonlinear function. %

\begin{table}[h!]
    \centering
    \caption{Values of prescribed parameter distributions for simulated models.}
    \label{tab:sim-prescr-values}
    \begin{tabular}{rcccccccc}
        & & & & & & & & \\
        & \multicolumn{3}{@{}c@{}}{Random array system} & & \multicolumn{4}{@{}c@{}}{Bridge truss system} \\
        \cmidrule{2-4} \cmidrule{6-9} 
       Parameter $\theta$ & $k$ [\si{N/m}] & $c$ [\si{Ns/m}] & $\kappa$ [\si{N/m^3}] & & $k$ [\si{N/m}] & $c$ [\si{Ns/m}] & $k_r$ [\si{N/rad}] & $\Phi$ [\si{\degree}] \\
       \midrule
       Mean $\mu_{\theta}$  & 200.0 & 0.1 & 1,000.0 & & 2,000.0 & 0.1 & 100.0 & 1.0 \\
       STD $\sigma_{\theta}$ & 10.0 & 5e-3 &	50.0 & & 100.0 & 0.005 &	5.0&	0.05\\
       \bottomrule
    \end{tabular}
\end{table}

The second source of error is in the form of measurement noise; the measured acceleration and force signals are corrupted with Gaussian white noise equivalent to a signal-to-noise ratio of 25. %

\section{Results}

For each system type, the offline training is performed on a smaller system (denoted the \emph{training set}), and online prediction is performed on a larger similar system and over a longer time window (denoted the \emph{testing set}). %
Details of the different model-system combinations for each prediction scheme are given in \Cref{tab:problem_assortment}. %
An important note is that the larger systems all contain the same \emph{form} of nonlinearity, but with varied underlying true parameters as shown in \Cref{tab:sim-prescr-values}. %

\begin{table}[h!]
    \caption{Problem testing assortment.}\label{tab:problem_assortment}
    \centering
    \begin{tabular*}{\textwidth}{@{\extracolsep\fill}ccccc}
    \toprule
             & \multicolumn{1}{@{}c@{}}{Offline training} & \multicolumn{3}{@{}c@{}}{Online prediction}\\
        \cmidrule{2-2}\cmidrule{3-5}
        System Type & & Models & System Size \& Window Lengths & Sparsities \\ 
        \midrule
        Sobol array & \begin{tabular}[c]{c}PGGNODE \\ 16 nodes \\ 4\si{s} \\ 75\%/87.5\% \end{tabular} & \begin{tabular}[c]{c}PGGNODE\\GEKF\end{tabular} & \begin{tabular}[c]{c}16 nodes - 4\si{s}\\32 nodes - 8\si{s}\\64 nodes - 8\si{s}\end{tabular} & \begin{tabular}[c]{c}75\%\\87.5\%\end{tabular} \\
        \hline
        Bridge truss & \begin{tabular}[c]{c}PGGNODE \\ 8m span \\ 2\si{s} \\ 75\%/87.5\% \end{tabular} & \begin{tabular}[c]{c}PGGNODE\\GEKF\end{tabular} & \begin{tabular}[c]{c}8m span - 2\si{s}\\16m span - 4\si{s}\\24m span - 4\si{s}\end{tabular} & \begin{tabular}[c]{c}75\%\si{s}\\87.5\%\si{s}\end{tabular} \\
    \bottomrule
    \end{tabular*}
\end{table}

The results are split into sections for offline and online prediction, where the former section shows the results of the initial training which additionally acts as a reference for schemes which might aim to simply utilise offline prediction. %
The latter section then shows having used the pretrained models to predict in an \emph{online} fashion, i.e. data is streamed directly to the prediction scheme. %

For assessment of the predicted structural dynamics, the \ac{nmse} is calculated, 
\begin{equation}
    \textrm{NMSE} = \frac{1}{2n_n}\sum_{j=1}^{n_n}\sum_{d\in\{x,y\}} \frac{||\hat{\mathbf{u}}_d-\mathbf{u}_d||^2}{{|\mathbf{u}_d|}}
    \label{eq:nmse}
\end{equation}
where $\mathbf{u}$ is the vector of the true values and $\bar{\mathbf{u}}$ is the norm of the true values. %

\subsection{Offline training}

In order to assess the capability of the PGGNODE to estimate the unobserved nodes on the graph-structure, this section begins with the results of the \emph{offline training} stage of the architecture. %
For all training and results, the state estimation is performed in a recurrent manner throughout the whole time window, with no batching. %

\subsubsection{Sobol array - cubic nonlinearity}

\begin{figure}[h!]
\centering
    \includegraphics[width=0.99\linewidth]{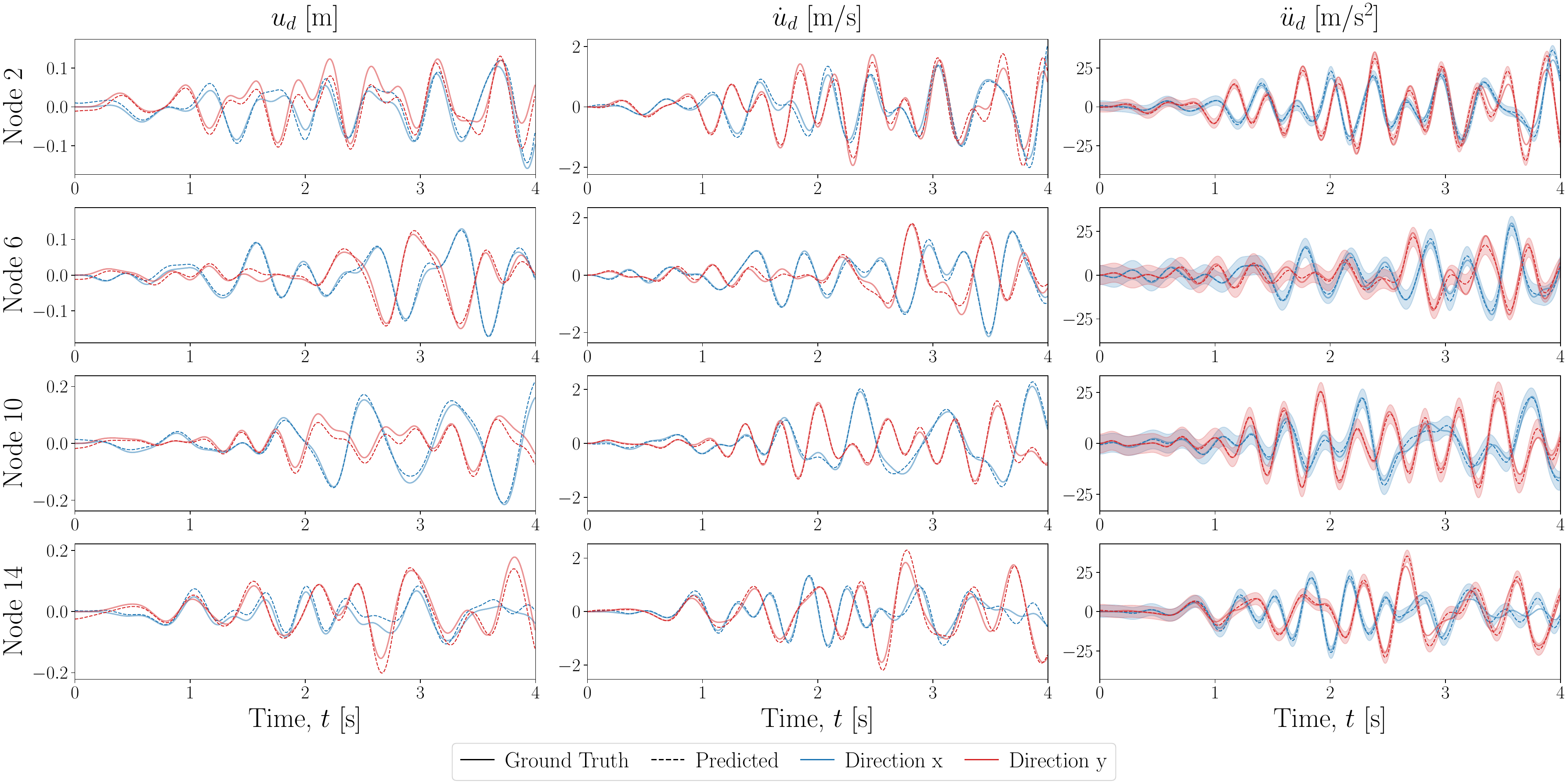}
    \caption{}
    \label{fig:rand_truss_gnode_preds_train}
\end{figure}

\Cref{fig:rand_truss_gnode_preds_train} shows the predicted and true signals of the offline training stage for the random truss array with cubic nonlinearity. %
The estimated signals match well with the ground truth, particularly in the acceleration and velocity estimations. %
There also appears to be an increasing bias in the state estimation, which is a common issue with evolutionary training methods with integration-based observations. %


\begin{figure}[h!]
    \includegraphics[width=0.99\linewidth]{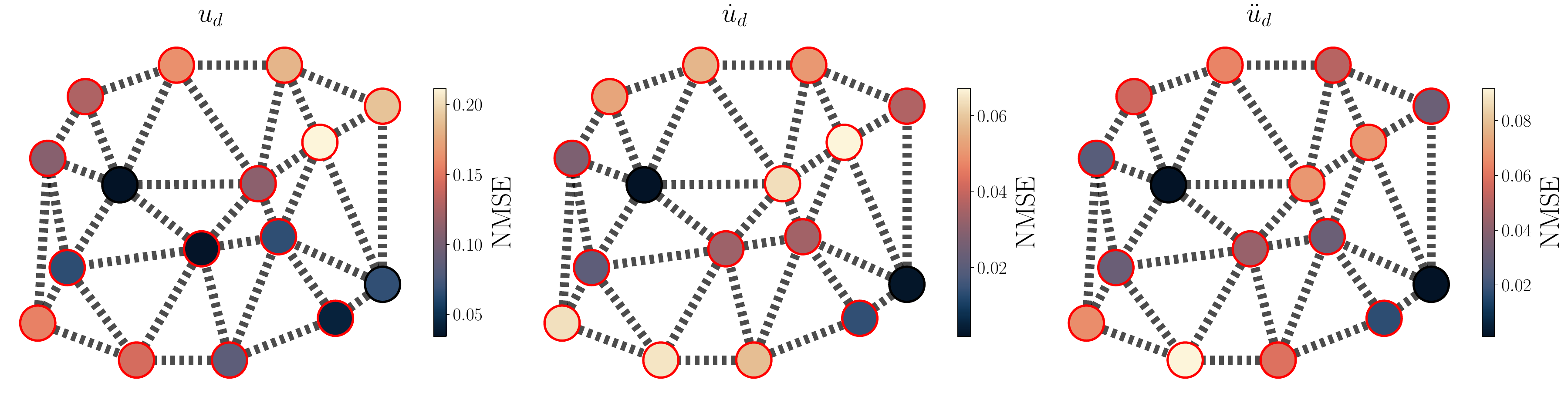}
    \caption{}
    \label{fig:rand_truss_gnode_mse_train}
    \caption{Normalised mean squared error (NMSE) per node of offline GNODE predictions over 2D random array training set.}
\end{figure}

To see any effects of topology on the estimation, \Cref{fig:rand_truss_gnode_mse_train} shows the \ac{nmse} at each node, for each dynamic variable. %
There appears to be no correlation between topology and quality of fit of the estimation to the ground truth. %
One potential influence would have been from boundary conditions, but there appears to be no correlation from this either. %
An additional potential cause may be the forcing location, but there also appears to be no effect from this. %


\subsubsection{Bridge truss - angular expansion nonlinearity}

\begin{figure}[h!]
    \centering
    \includegraphics[width=0.99\linewidth]{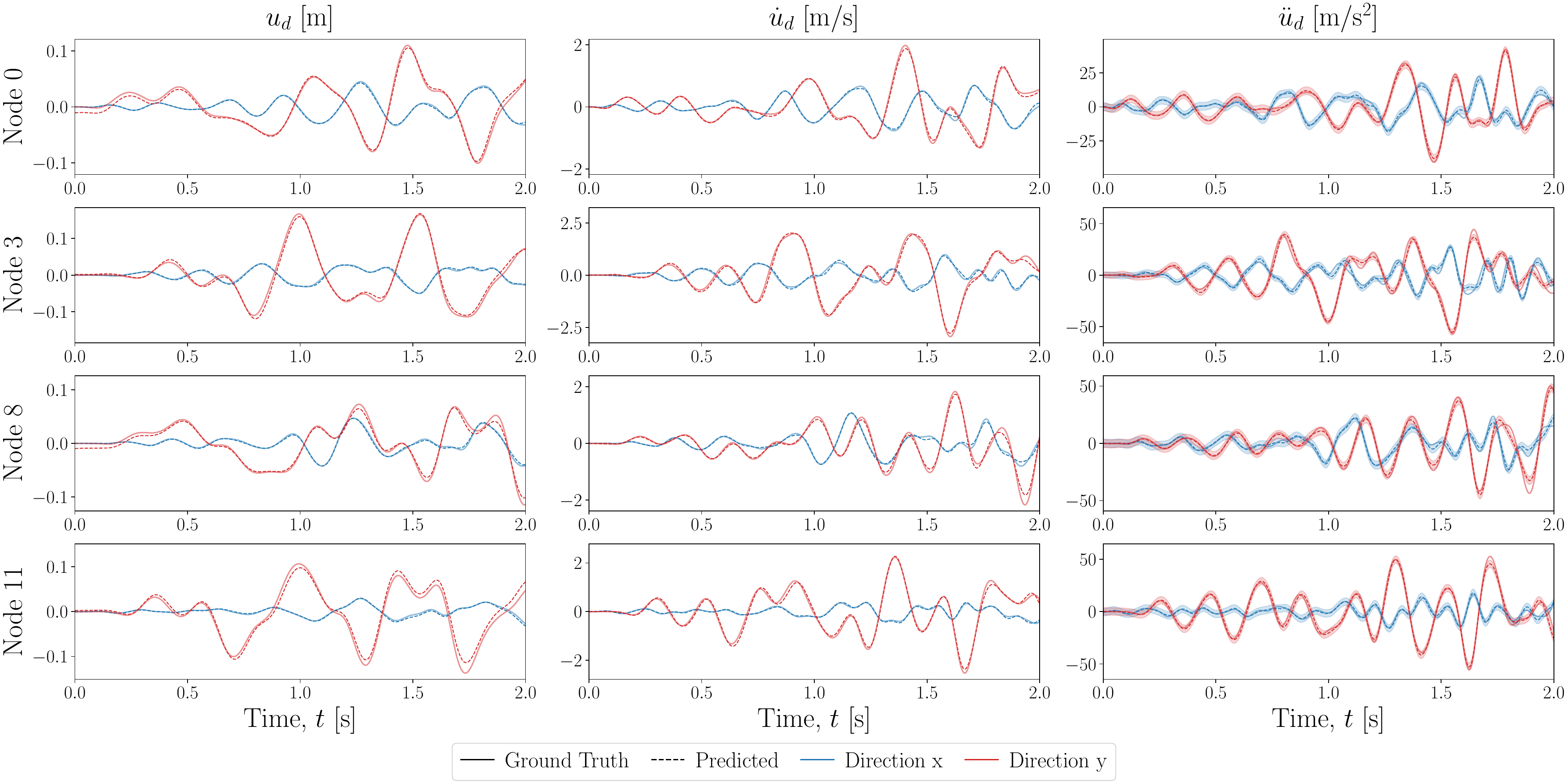}
    \caption{Offline GNODE predictions for the bridge truss system over the training dataset, illustrating the agreement between predicted and reference responses at selected measured and unmeasured nodes.}
    \label{fig:bridge_truss_gnode_preds_train}
\end{figure}

The results of the estimated signals for a selection of unmeasured nodes is shown in \Cref{fig:bridge_truss_gnode_preds_train}, where, as with the random array, the estimated signals match well with the ground truth. %
Compared to the random array, there appears to be less error in the displacement estimation, which is likely due to the higher stiffness of the system reducing integration drift. %

\begin{figure}[h!]
    \centering
    \includegraphics[width=0.66\linewidth]{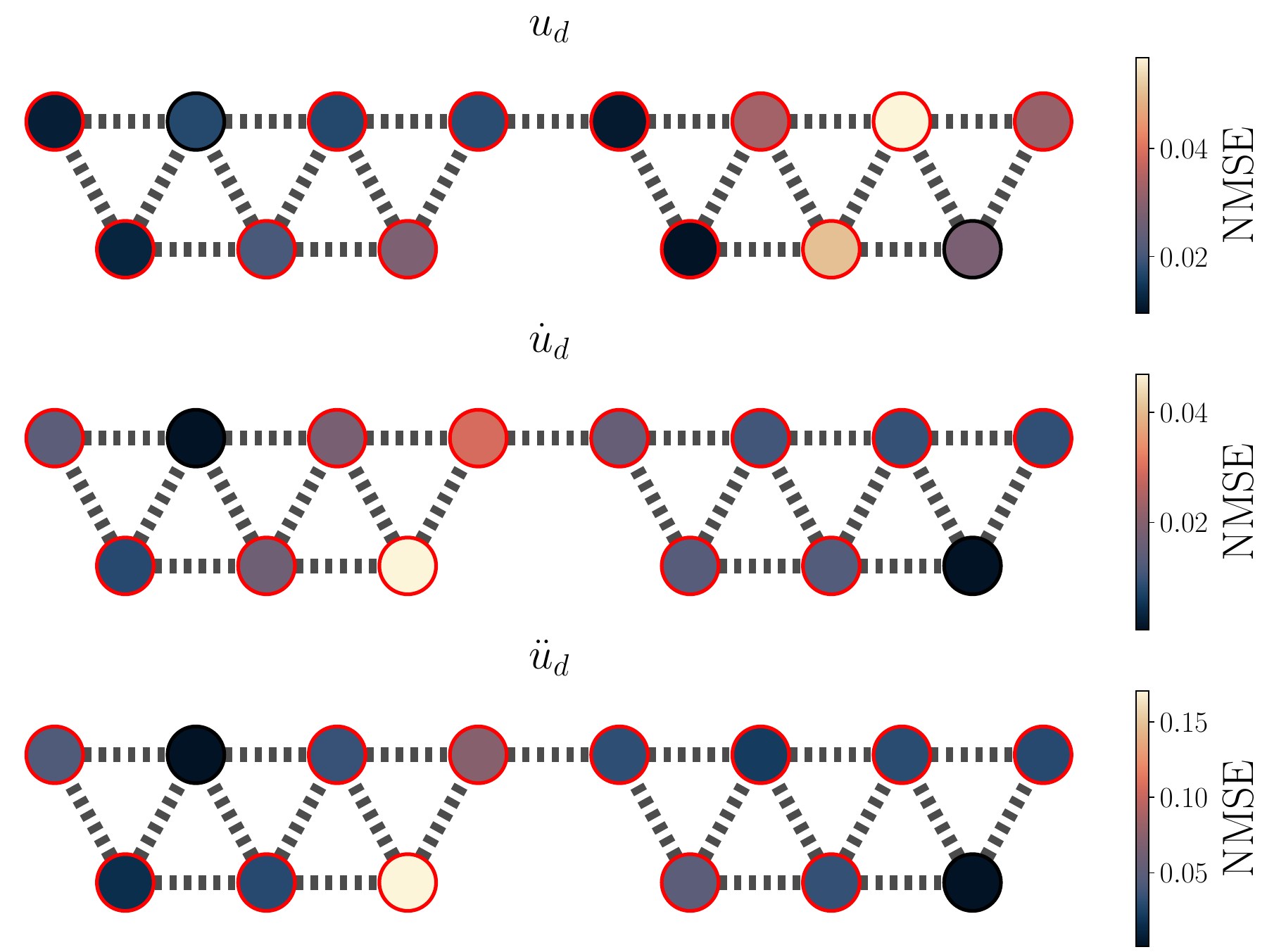}
    \caption{Normalised mean squared error (NMSE) per node of offline GNODE predictions over 2D bridge truss training set.}
    \label{fig:bridge_truss_gnode_mse_train}
\end{figure}

The \ac{nmse} for each node and the dynamic variable for the bridge truss is shown in \Cref{fig:bridge_truss_gnode_mse_train}, where again there is no strong effect of topology on the accuracy of the prediction. %


\subsection{Online virtual sensing}
Following the offline training of the systems, \emph{online sensing} is performed with different schemes. %
The first scheme is to directly employ the \ac{pggnode} to attempt to estimate the response of the structure using the measured force. %
The second scheme is to use the \ac{gkf} where the underlying transition model is performed by the pre-trained \ac{pggnode}. %
For plotting of response signal estimations, the signals are bounded by a shaded area representing the 95\% confidence interval, calculated as $\mu \pm 2\sigma$. %
For the \ac{pggnode} the confidence interval is given for the acceleration, where $\sigma_{\ddot{u}}$ is estimated from \Cref{eq:residual_var}. %
For estimation from the \ac{gkf} models, the state signal variances are extracted from the diagonal of $\Sigma_{t|t}$ returned from the filtering scheme. %
The estimated variance of the observation signal (acceleration) is calculated as,
\begin{equation}
    \Sigma_{o} = \mathbf{H}_{t|t} \Sigma_{t|t} \mathbf{H}_{t|t}^T + \mathbf{R}_t
\end{equation}
where $\mathbf{H}_{t|t}=\nabla_{\mathbf{Y}}(f_{\textrm{o}}(\mathcal{G}_{t|t}))$ is the Jacobian of the observation function at the posterior estimate. %

\subsubsection{Random truss - cubic nonlinearity}


\paragraph{GNODE model prediction}

The response signals for four unmeasured nodes, estimed using solely using the \ac{pggnode} model are shown in \Cref{fig:rand_truss_gnode_preds_train}, for the latter four seconds of the signal. %
Compared to the offline prediction, these signals are much less accurate to the ground truth. %
The predictions offer a smooth response estimation, indicating that the estimated form of the function of the nonlinearity to be smooth. %
However, there is significant phase drift and amplitude discrepancies, indicating that the learnt model is not adequately generalised to structures with variable structural parameters. %


\begin{figure}[h!]
    \centering
    \includegraphics[width=0.99\linewidth]{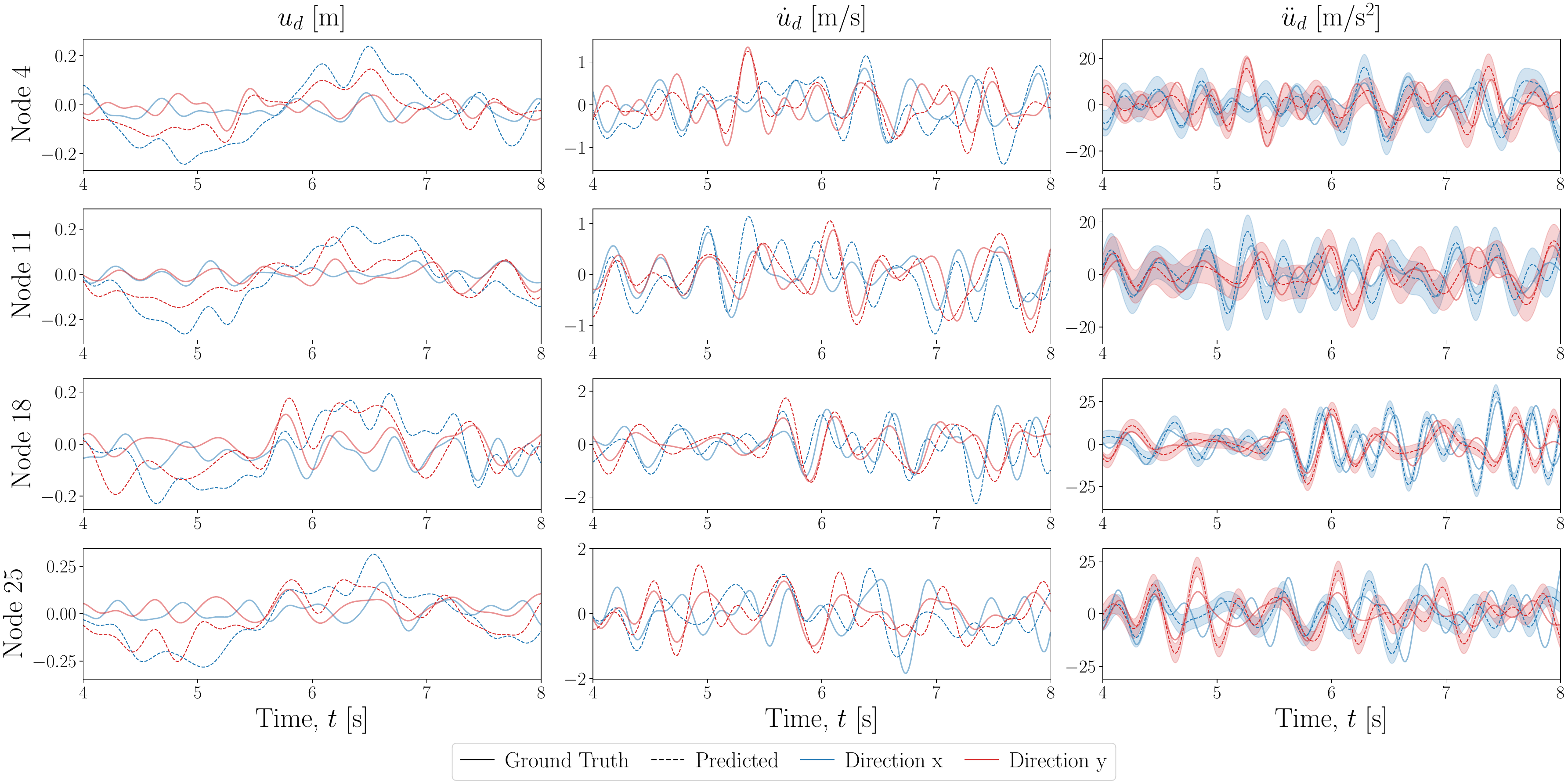}
    \caption{Sobol array online GNODE predictions over testing set}
    \label{fig:rand_truss_gnode_preds_test}
\end{figure}

\begin{figure}[h!]
    \centering
    \includegraphics[width=0.99\linewidth]{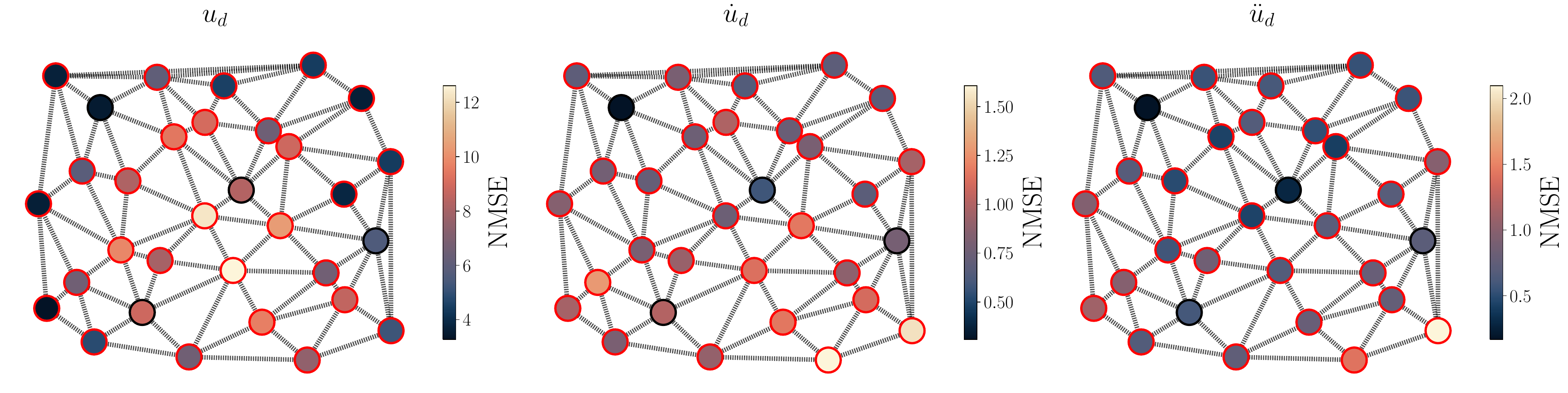}
    \caption{Normalised mean squared error (NMSE) per node of online GNODE predictions over 2D sobol array testing set.}
    \label{fig:rand_truss_gnode_mse_test}
\end{figure}

\Cref{fig:rand_truss_gnode_mse_test} shows the \ac{nmse} at each node for the predicted response using the \ac{pggnode}. %
For acceleration and velocity estimation, there does not appear to be any correlation between topology and prediction accuracy, however, the appears to be larger error towards the "left" part of the structure for the displacement estimation. %
As the topology, forcing and boundary conditions are symmetric, this is likely to be a non-deterministic grouping of integration error. %


\paragraph{GEKF model prediction}

\begin{figure}[h!]
    \centering
    \includegraphics[width=0.99\linewidth]{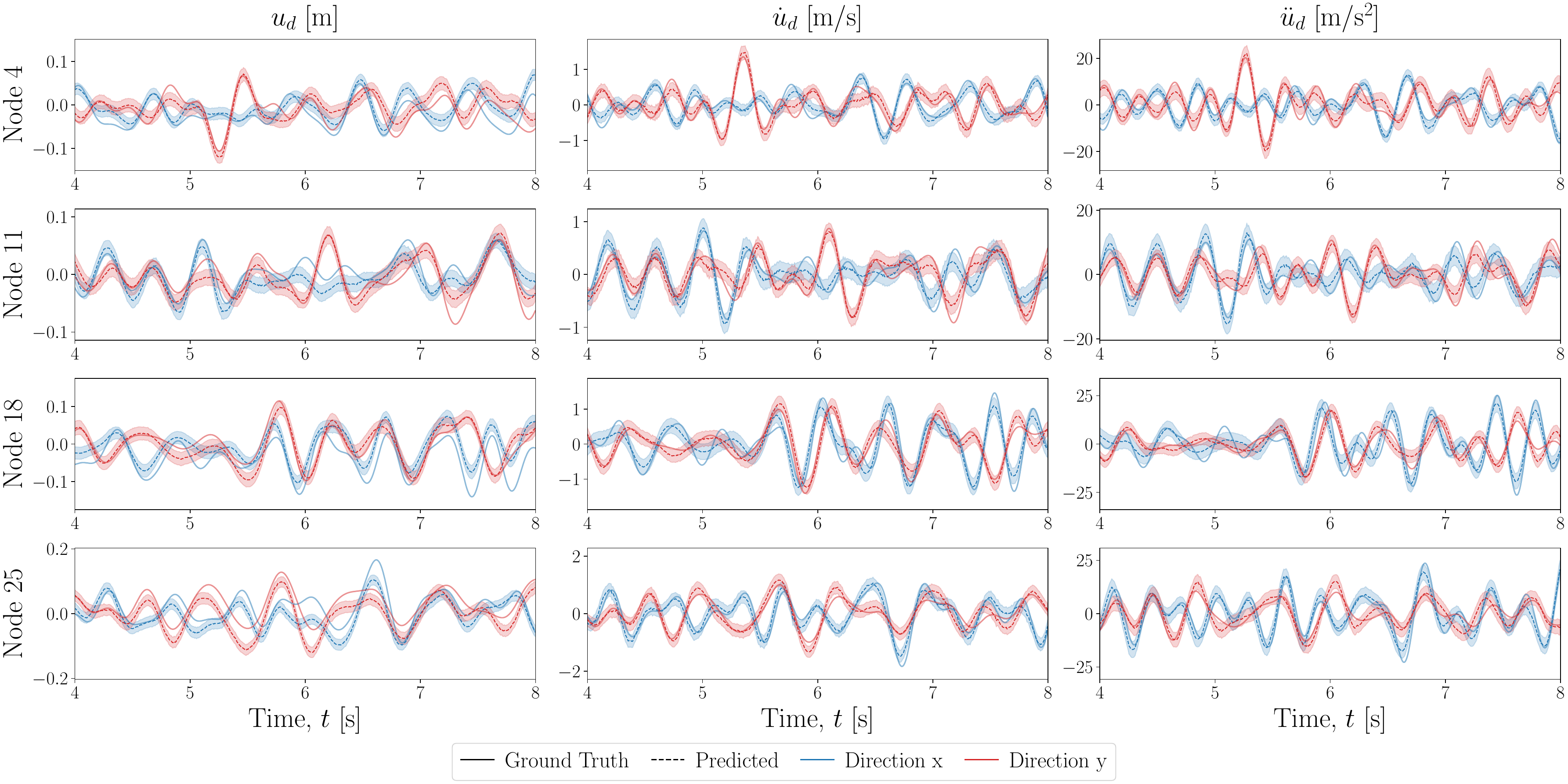}
    \caption{Sobol array online GEKF predictions over testing set}
    \label{fig:rand_truss_gekf_preds_test}
\end{figure}

\Cref{fig:rand_truss_gekf_preds_test} shows the response estimation an in online virtual sensing context using the \ac{gkf} model, which, compared to the \ac{pggnode} prediction, is much more accurate to the ground truth. %
There appears to be a constant bias error for the displacement prediction signal, similar to the offline training results, which is again a common issue with integration-based estimation schemes. %
The posterior estimate of the covariance show a reasonable uncertainty in the prediction, and is consistent over all the plotted signals. %


\begin{figure}[h!]
    \centering
    \includegraphics[width=0.99\linewidth]{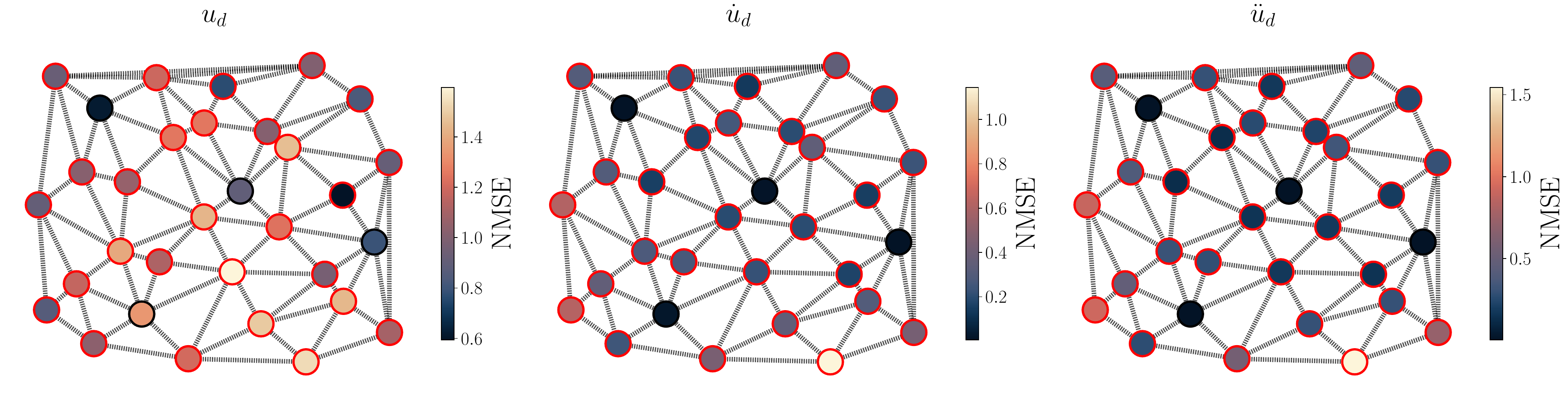}
    \caption{Normalised mean squared error (NMSE) per node of online GEKF predictions over 2D random array testing set.}
    \label{fig:rand_truss_gekf_mse_test}
\end{figure}

The \ac{nmse} at each node for the \ac{gkf} prediction is shown in \Cref{fig:rand_truss_gekf_mse_test}. %
The first notable observation is the much lower error for all variables compared to the \ac{pggnode} prediction. %
Though the displacement estimation is has lower error, it does have again the `grouping' of the error to an area of the graph. %
However, this time it is in a different location, indicating that this grouping is nondeterministic to the structural topology and boundary conditions. %


\subsubsection{Bridge truss - angular expansion nonlinearity}

\paragraph{GNODE model prediction}

\begin{figure}[h!]
    \centering
    \includegraphics[width=0.99\linewidth]{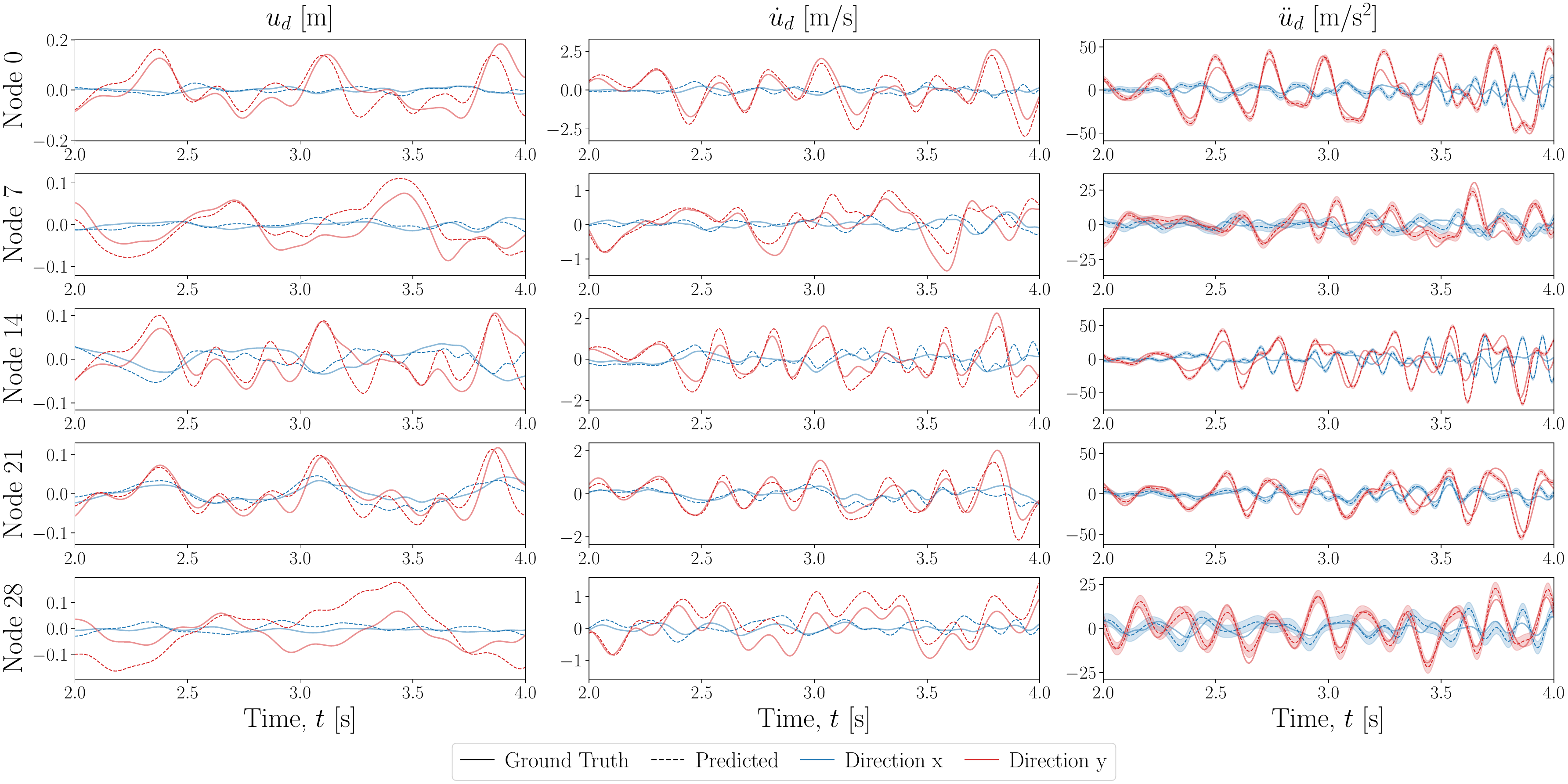}
    \caption{Bridge truss online GNODE predictions over testing set}
    \label{fig:bridge_truss_gnode_preds_test}
\end{figure}

The results of the online prediction using the \ac{pggnode} model on the testing set bridge truss structure are shown in \Cref{fig:bridge_truss_gnode_preds_test}. %
Compared to the online \ac{pggnode} predictions for the random array, the bridge truss system appears to have better accuracy to the ground truth. %
This could be explained together with the lesser displacement error of the offline training, as a result of the overall stiffer system resulting in improved generalisation for locally variational structural parameters. %


\begin{figure}[h!]
    \centering
    \includegraphics[width=0.99\linewidth]{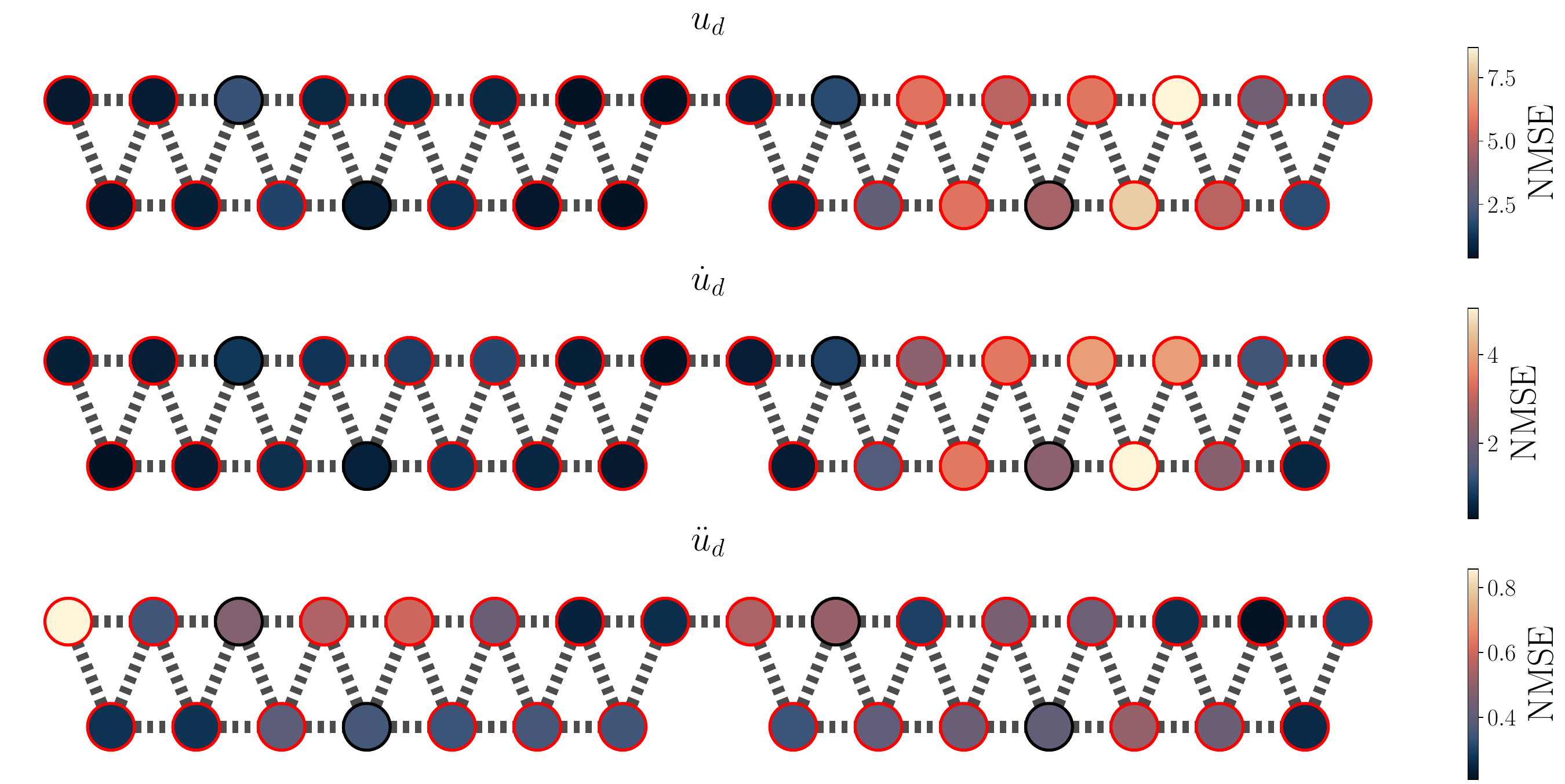}
    \caption{Normalised mean squared error (NMSE) per node of online GNODE predictions over 2D bridge trust testing set.}
    \label{fig:bridge_truss_gnode_mse_test}
\end{figure}

\Cref{fig:bridge_truss_gnode_mse_test} shows the \ac{nmse} at each node for the online predictions of the bridge truss testing structure using \ac{pggnode}. %
Similar to the random array, there appears to be a grouping of nodes with lower accuracy, though this time this inaccuracy is for all dynamic variables, not solely displacement. %
This highlights further the indetermenistic nature of the error fo the predictions using this scheme. %


\paragraph{GEKF model prediction}

\begin{figure}[h!]
    \centering
    \includegraphics[width=0.99\linewidth]{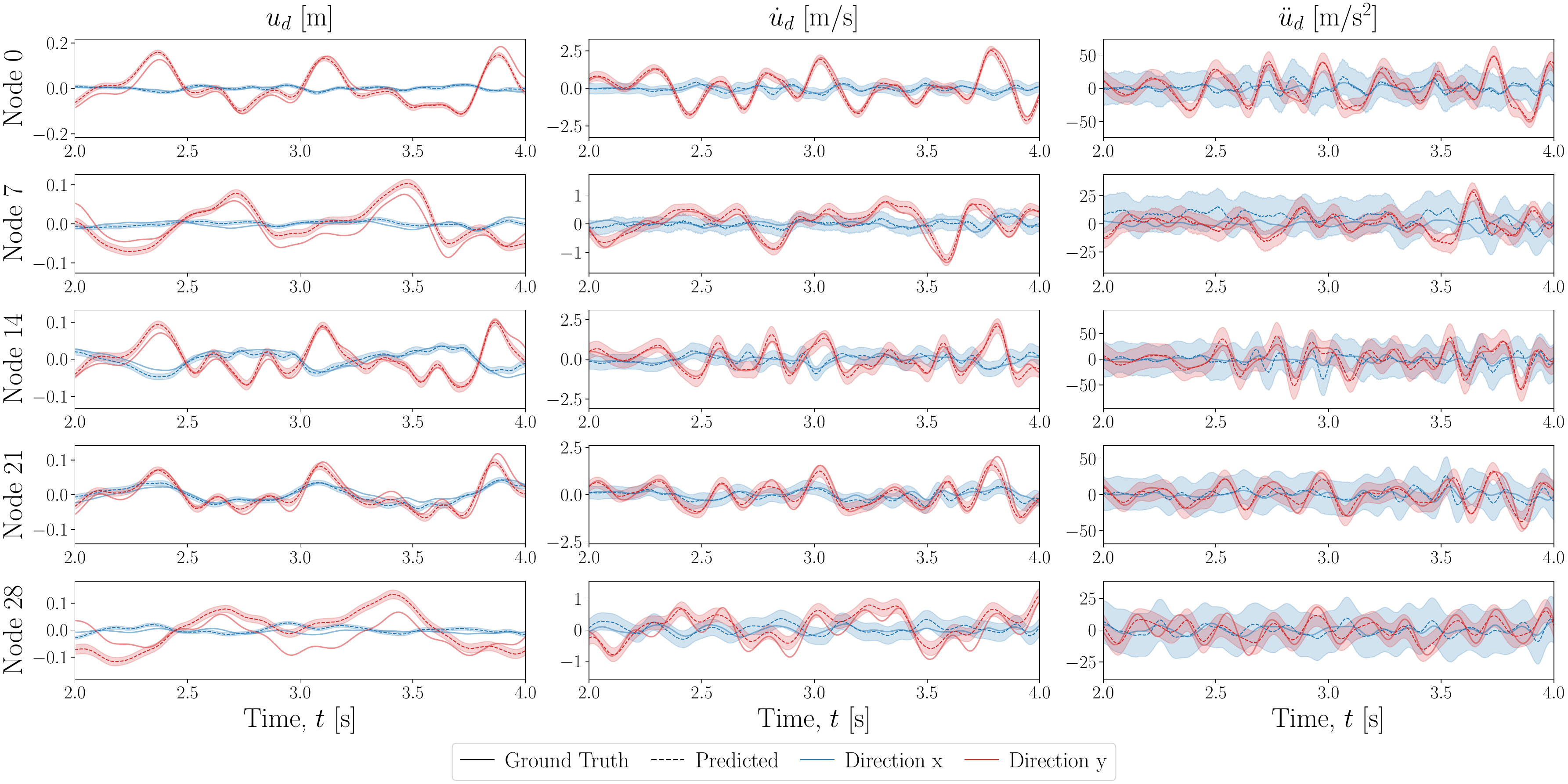}
    \caption{Bridge truss online GEKF predictions over testing set}
    \label{fig:bridge_truss_gekf_preds_test}
\end{figure}

The results of the online prediction using the \ac{gkf} model on the testing set bridge truss structure are shown in \Cref{fig:bridge_truss_gekf_preds_test}. %
Compared to the online \ac{pggnode} predictions, the \ac{gkf} appears to have better accuracy to the ground truth. %
There is once again the issue of constant bias for the displacement prediction. %


\begin{figure}[h!]
    \centering
    \includegraphics[width=0.99\linewidth]{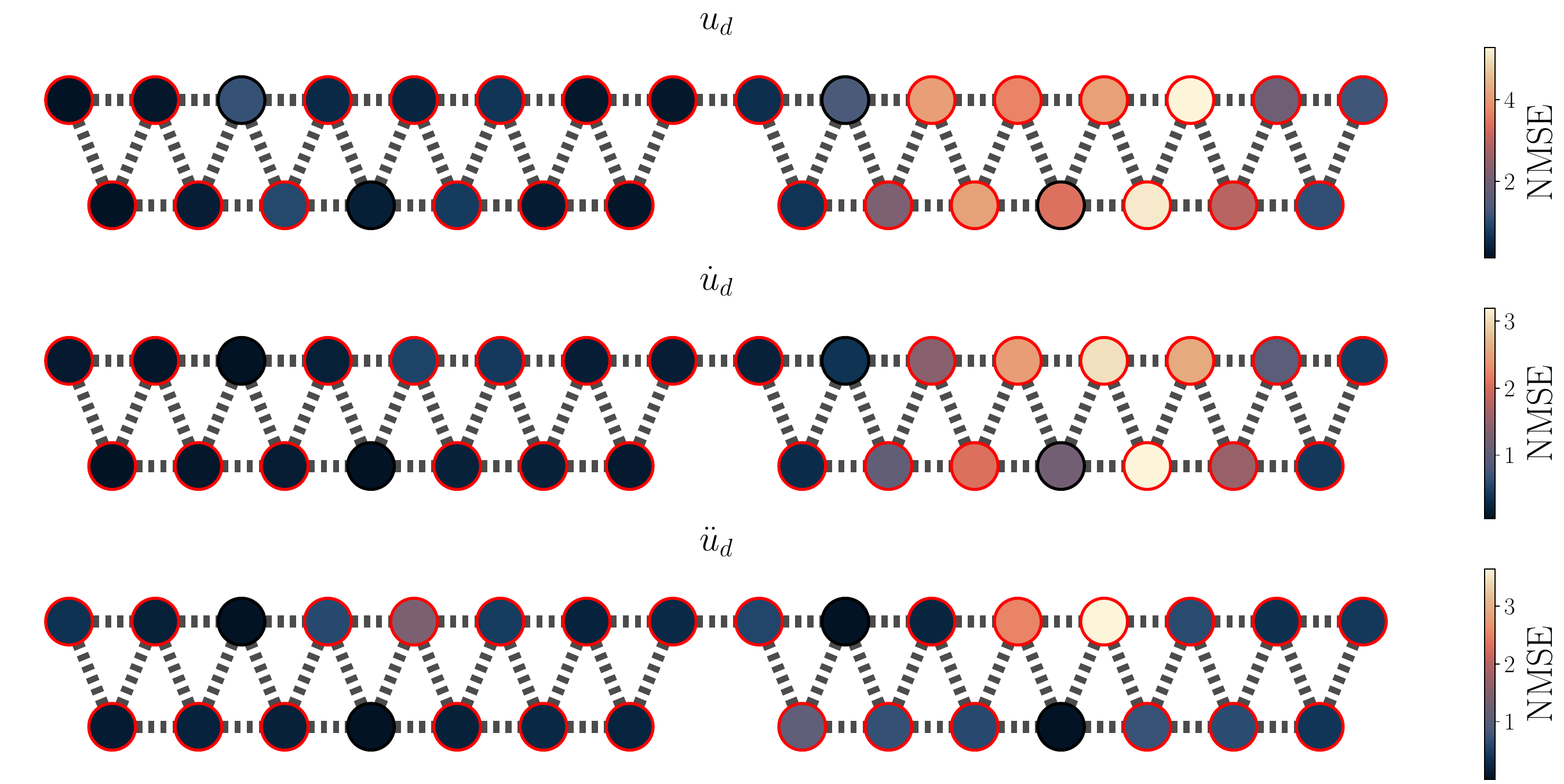}
    \caption{Normalised mean squared error (NMSE) per node of online GEKF predictions over 2D bridge truss testing set.}
    \label{fig:bridge_truss_gekf_mse_test}
\end{figure}

The per-node \ac{nmse} for the \ac{gkf} prediction is shown in \Cref{fig:bridge_truss_gekf_mse_test}, where the local clustering of prediction error appears to be lessened, though is still possibly apparent for the displacement predictions. %
For the entire topology, the \ac{gkf} predictions have much lower error compared to the \ac{pggnode} for online prediction. %


\begin{figure}[h!]
    \centering
    \includegraphics[width=0.95\linewidth]{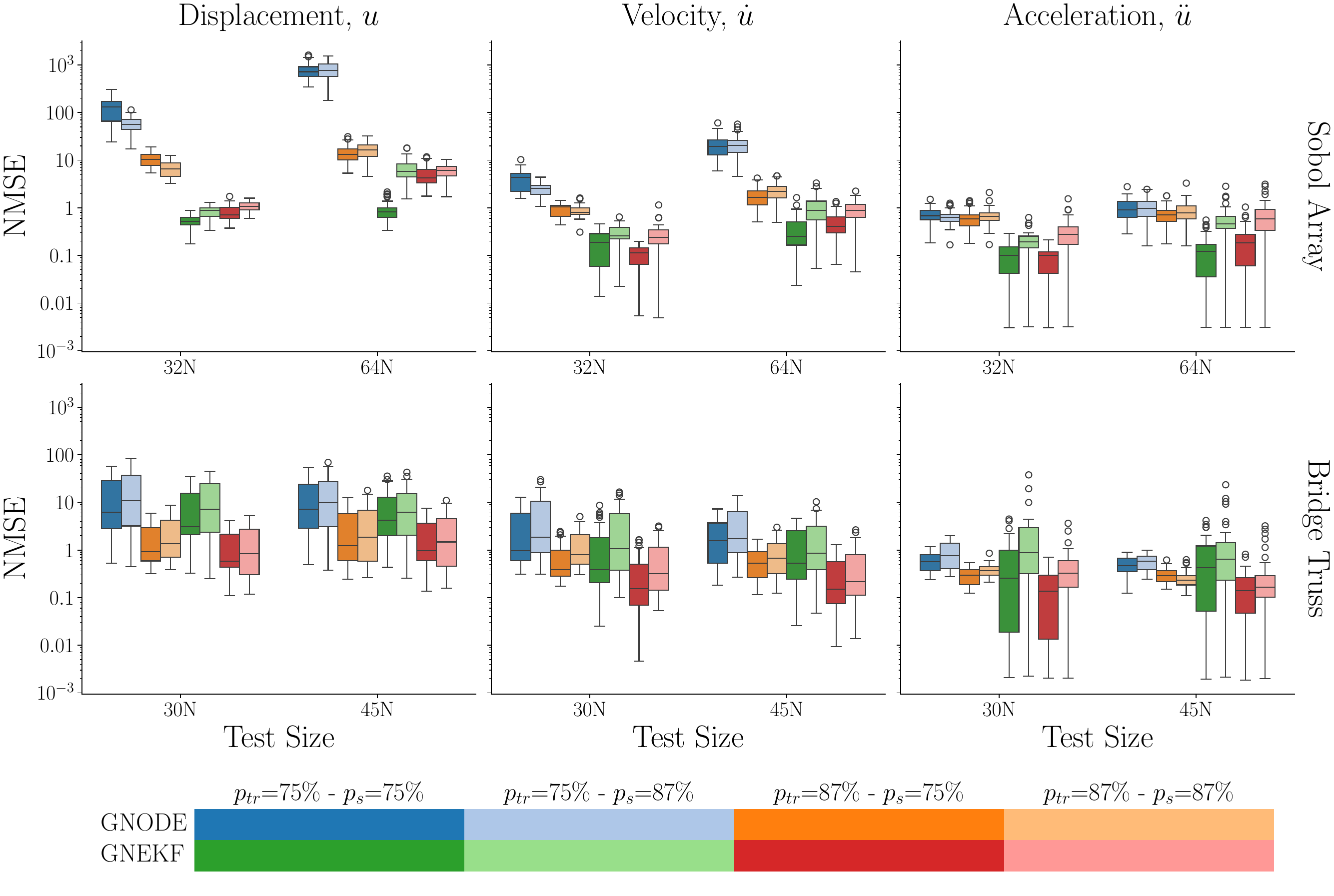}
    \caption{Comparison of averaged NMSE values for different combinations of training sparsity, testing sparsity and prediction model for the Sobol array and bridge truss systems.}
    \label{fig:sobol_nmse_comp}
\end{figure}

A summary of the total \ac{nmse} over the entire structure is given in \Cref{tab:nmse-summary}, where each variable is calculated using \Cref{eq:nmse} independently. %
It can be seen from the table that the \ac{gkf} consistently reduces the error in the prediction of response of the structure, given a noisy measured input. %

\begin{table}[h!]
    \caption{Summary of NMSE of the case studies.}\label{tab:nmse-summary}
    \centering
    \begin{tabular*}{\textwidth}{@{\extracolsep\fill}ccccccc}
    \toprule
             & \multicolumn{3}{@{}c@{}}{PGGNODE} & \multicolumn{3}{@{}c@{}}{PGGNODE-GEKF}\\
        \cmidrule{2-4}\cmidrule{5-7}
        System & $\mathbf{u}$ & $\dot{\mathbf{u}}$ & $\ddot{\mathbf{u}}$ & $\mathbf{u}$ & $\dot{\mathbf{u}}$ & $\ddot{\mathbf{u}}$ \\ 
        \midrule
        \midrule
        16-node random array & 4.077 & 0.020 & 0.013 & 3.463 & 0.034 & 0.023     \\
        \hline
        32-node random array & 24.538 & 2.644 & 2.730 & 0.298 & 0.016 & 0.016      \\
        \hline
        \hline
        8m wide bridge & 0.088 & 0.052 & 0.075 & 0.051 & 0.012 & 0.026      \\
        \hline
        16m wide bridge & 30.199 & 0.377 & 0.316 & 0.758 & 0.027 & 0.052      \\
    \bottomrule
    \end{tabular*}
\end{table}

\pagebreak

\section{Discussion}


The first stage of the proposed scheme involves offline training of the \ac{pggnode} model, which by itself additionally acts as an approach for offline virtual sensing. %
The accuracy of the virtual sensing scheme for recovering the response at unmeasured nodes is highlighted in \Cref{fig:rand_truss_gnode_preds_train,fig:rand_truss_gnode_mse_train,fig:bridge_truss_gnode_preds_train,fig:bridge_truss_gnode_mse_train}, where it shows the inductive biases provide a powerful model for such a scheme. %
However, such a model appears to not be well generalised for structures with uncertainty in local properties, as indicated by the lower performance when estimating the response of a different larger, but topologically and functionally similar, structure. %
The smoothness of the prediction, indicates that the estimated function is adequately smooth, but the model form assumes constant model parameter; i.e. it estimates the nonlinearity to have the same stiffness or clearance angle everywhere in the structure. %

By implementing the learnt \ac{pggnode} model into a \ac{gkf} framework, online predictions on unseen structures are significantly improved. %
As seen in the \ac{pggnode} predictions, the model suffers from unknown variation in structural properties, which reduces generalisation over structural topology. %
Furthermore, although the predicted uncertainty is larger, the filtering scheme provides an improved and more appropriate estimate. %
Not only does the scheme provide estimates for all dynamic variables, the uncertainty bounds enevelope the true responses, indicating a better match. %
This is intuitive, as the uncertainty is estimated directly from comparisons between predictions and data, thus capture model form and noise uncertainty, as opposed to with the \ac{pggnode} scheme where they are estimated from equations which assume no model uncertainty. %

As the \ac{gkf} model uses the extended Kalman filter scheme, it can be deduced that the \ac{pggnode} is well generalised to the model \emph{form} (reminder; it assumes the local nonlinearities to have the same form and parameter). %
The filtering schemes assumes Gaussian error which does not handle discrete errors well, indicating that the representative equation of the learned model is inaccurate in the form of a normally-distributed error. %
However, this also highlights a potential limitation; that the underlying uncertainty which reduces generalisation follows a Gaussian distribution, though this could be tackled by employing \ac{kf} schemes adapted to handle non-Gaussian distributions, such as unscented or particle filtering schemes. %

An additional limitation of the proposed scheme is that the offline training requires the known initial position. %
As the nature of the scheme is to learn a generalised model for structures of similar types, but varying topologies, this limitation may not be excessively restricting, as the learning stage can be performed in a controlled or experimental context, where initial states are known. %
For scenarios where initial states are difficult to obtain, such as a population of in-situ structures with no rest, then one would need to train a Graph-Neural Extended Kalman Filter with a long time window, to overcome initial state error. %
The longer required time window, combined with the additional training model complexity, would present a difficult technical challenge for training and is beyond the scope of this work. %

A final consideration of the architecture is that there is an assumption of global nonlinearities, i.e.\ that the nonlinearities exist and are of the same form over the entire structure. %
Further work is required to account for such problems, where the achitecture can be adapted to include local estimation of the presence, or form, of nonlinearities. %
This could be done with layered \acp{gnn}, where an additional layer is used as a `switching' model for the nonlinearity. %
An alternative approach is to adapt the prescription of the \ac{gkf} by Alippi \& Zambon \cite{alippi2023graph}, where the adjacency matrix is an additional accountable uncertainty, which determines the time-variant existence of edge connections representing nonlinearities. %

\section{Conclusion}

This study introduced the Physics-Guided Graph Neural ODE (PiGGO) framework, which combines a graph-based, physics-informed state evolution model with extended Kalman filtering for online virtual sensing of nonlinear structural systems. %
The proposed approach leverages inductive biases derived from structural mechanics to constrain the learning of nonlinear dynamics, enabling robust performance under sparse sensing and model-form uncertainty, where only nominal system parameters are assumed \emph{a priori}. %

The numerical investigations demonstrate that the PiGGO--EKF scheme achieves accurate and stable state estimation across structurally similar systems with varying properties, outperforming purely data-driven and open-loop graph neural models in terms of generalisation and robustness. %
In addition, the probabilistic filtering formulation provides uncertainty-aware predictions, with posterior covariance estimates that meaningfully capture the variability in the true system response. %

Overall, the framework offers a unified approach for integrating physics-guided learning and Bayesian filtering within a graph-based representation of structural systems. %
Future work will focus on extending the methodology to more complex nonlinearities, including localised and non-smooth behaviours, as well as exploring adaptive graph representations and fully integrated end-to-end learning of the filtering process.

\section*{Supplementary information}

All the code and data for this article is available open access at a Github repository available at \href{https://github.com/MarcusHA94/struct-pggnode}{https://github.com/MarcusHA94/struct-pggnode}. %

\section*{Acknowledgements}

The authors gratefully acknowledge the funding from the State Secretariat for Education, Research, and Innovation (SERI) as matching funding for the Horizon Europe project ‘ReCharged - Climate-aware Resilience for Sustainable Critical and interdependent Infrastructure Systems enhanced by emerging Digital Technologies’ (grant agreement No: 101086413), and funding from the French National Research Agency (ANR PRCI Grant No. 266157) and the Swiss National Science Foundation (Grant No. 200021L\_212718) for the MISTERY project.

\section*{Declarations}

The authors declare no conflicts of interest.

\begin{appendices}

    \section{Corotational Kinematics Calculations}
    \label{app:corot_kinematics}
    
    
    \begin{itemize}
        \item $\mathbf{p}_{0,k} \in \mathbb{R}^2$: The nominal position of node $k$ in $x$ and $y$ directions.
        \item $\mathbf{u}_k(t) \in \mathbb{R}^2$: The displacement of node $k$ in $x$ and $y$ directions.
        \item $\mathbf{p}_k(t) = \mathbf{p}_{0,k} + \mathbf{u}_k$: The current position of node $k$.
        \item $\mathbf{v}_k(t) = \frac{d\mathbf{p}_k}{dt}$: The velocity of node $k$.
        \item $\mathbf{p}_{a,j} \in \mathbb{R}^2$: The position of a fixed anchor point for a boundary node $j$.
        \item $L_{0,ij}$: The rest length of the edge.
    \end{itemize}
    The current vector for edge $ij$ is
    \begin{equation}
        \mathbf{d}_{ij} = \mathbf{p}_j - \mathbf{p}_i
    \end{equation}
    And for self-loop edges (boundary edges), with anchor points $\mathbf{p}_{a,i}$
    \begin{equation}
        \mathbf{d}_{ii} = \mathbf{p}_i - \mathbf{p}_{a,i}
    \end{equation}
    Then, the edge length and direction are
    \begin{equation}
        l_{ij} = ||\mathbf{d}_{ij}||, \qquad \hat{\mathbf{n}}_{ij} = \frac{\mathbf{d}_{ij}}{l_{ij}}
    \end{equation}
    Finally, the edge extension, rate, and angle are,
    \begin{equation}
        \epsilon_{ij} = l_{ij} - L_{0,ij}, \qquad \dot{\epsilon}_{ij} = (\mathbf{v}_j - \mathbf{v}_i) \cdot \hat{\mathbf{n}}_{ij}, \qquad \theta_{ij} = \operatorname{atan2}((\hat{\mathbf{n}}_{ij})_y, (\hat{\mathbf{n}}_{ij})_x)
    \end{equation}

\end{appendices}


\bibliographystyle{elsarticle-num}
\bibliography{gnodekf_bib}

\end{document}